\RequirePackage{snapshot}
\documentclass[letterpaper, 10 pt, conference]{ieeeconf}  
\IEEEoverridecommandlockouts                              

\usepackage{amsfonts,amsmath,amssymb,mathtools}
\usepackage[romanian]{babel}
\usepackage{combelow}
\usepackage{bigints}
\usepackage{hhline}
\usepackage{marvosym} 

\usepackage{balance}

\renewcommand{\thepage}{}

\usepackage{algorithm}
\usepackage{algpseudocode}
\makeatletter
\let\OldStatex\Statex
\renewcommand{\Statex}[1][0]{%
  \setlength\@tempdima{\algorithmicindent}%
  \OldStatex\hskip\dimexpr#1\@tempdima\relax}
\makeatother

\algnewcommand\AND{~\textbf{and}~}
\algnewcommand\OR{~\textbf{or}~}
\algnewcommand\CONTINUE{~\textbf{continue}~}
\DeclareMathOperator*{\argmin}{arg\,min} 

\makeatletter
\algnewcommand{\LineComment}[1]{\Statex \hskip\ALG@thistlm \(\triangleright\)
  #1}
\makeatother

\usepackage{graphicx}
\usepackage{booktabs, multicol, multirow}
\usepackage{indentfirst}
\usepackage{subfigure}
\usepackage[font=footnotesize]{caption}
\usepackage[percent]{overpic}

\usepackage{marginnote}
\usepackage[usenames,dvipsnames]{color}
\usepackage[table]{xcolor}    
\definecolor{fullred}{rgb}{0.85,.0,.1} 
\definecolor{navyblue}{rgb}{.0,.0,.5}
\definecolor{bleudefrance}{rgb}{0.19, 0.55, 0.91}
\definecolor{bluegray}{rgb}{0.18, 0.36, 0.6}
\definecolor{lightgray}{rgb}{0.95, 0.95, 0.95}
\definecolor{white}{rgb}{1.0, 1.0, 1.0}

\usepackage{array}
\newcolumntype{+}{>{\global\let\currentrowstyle\relax}}
\newcolumntype{^}{>{\currentrowstyle}}

\newcolumntype{C}[1]{>{\centering\arraybackslash}p{#1}}

\makeatletter
\let\NAT@parse\undefined
\makeatother

\usepackage{url}
\usepackage[pagebackref=true,bookmarks=true, colorlinks=true, urlcolor=Blue]{hyperref}
\usepackage{multicol}

\newcommand{\ME}{Sudeep Pillai}
\newcommand{\PAPERAUTHORS}{Sudeep Pillai, Rare\cb{s} Ambru\cb{s}, Adrien Gaidon}
\newcommand{\PAPERTITLE}{SuperDepth: Self-Supervised, Super-Resolved Monocular Depth Estimation}
\newcommand{\PAPERKEYWORDS}{Machine Learning; Deep Learning; Computer Vision; Perception; Stereopsis}

\hypersetup{pdfauthor={\ME},%
            pdftitle={\PAPERTITLE},%
            pdfsubject={\PAPERTITLE},%
            pdfkeywords={\PAPERKEYWORDS},%
            pdfproducer={LaTeX},%
            pdfcreator={pdfLaTeX}
}

\begin{document}

\title{\LARGE \PAPERTITLE\vspace{-2ex}}
\author{\PAPERAUTHORS{}~\\
Toyota Research Institute (TRI)
\thanks{The authors are with the Toyota Research Institute (TRI) 4440 El Camino Real, Los Altos, CA 94022 USA and can be reached via email at {\tt\small firstname.lastname@tri.global}}
}

\maketitle\vspace{-6mm}
\thispagestyle{empty}
\pagestyle{empty}

\begin{abstract}
Recent techniques in self-supervised monocular depth estimation
are approaching the performance of supervised methods, but operate in low resolution only. We show that high resolution is key towards high-fidelity self-supervised monocular depth prediction.
Inspired by recent deep learning methods for Single-Image Super-Resolution, we propose a \textit{sub-pixel convolutional layer extension} for depth super-resolution that accurately synthesizes high-resolution disparities from their corresponding low-resolution convolutional features.
In addition, we introduce a \textit{differentiable flip-augmentation layer} that accurately fuses predictions from the image and its horizontally flipped version, reducing the effect of left and right shadow regions generated in the disparity map due to occlusions.
Both contributions provide significant performance gains over the state-of-the-art in self-supervised depth and pose estimation on the public KITTI benchmark. A video of our approach can be found at \href{https://youtu.be/jKNgBeBMx0I}{https://youtu.be/jKNgBeBMx0I}.



\end{abstract}

\IEEEpeerreviewmaketitle


\section{Introduction}
\label{sec-introduction}
Robots need the ability to simultaneously infer the 3D structure of a scene and estimate their ego-motion to enable autonomous operation. Recent advances in Convolutional Neural Networks (CNNs), especially for depth and pose estimation~\cite{ummenhofer2017demon,wang2018learning,kuznietsov2017semi,eigen2014depth} from a monocular camera have dramatically shifted the landscape of single-image 3D reconstruction. These methods cast monocular depth estimation as a supervised or semi-supervised regression problem, and require large volumes of ground truth depth and pose measurements that are sometimes difficult to obtain. On the other hand, self-supervised methods in depth and pose estimation~\cite{garg2016unsupervised,godard2017unsupervised,zhou2017unsupervised} alleviate the need for ground truth labels and provide a mechanism to learn these latent variables by incorporating geometric and temporal constraints to effectively infer the structure of the 3D scene.

 Recent works~\cite{godard2017unsupervised,zhou2017unsupervised,godard2018digging} in \textit{self-supervised} depth estimation are limited to training in lower-resolution regimes due to the large memory requirements of the model and their corresponding self-supervised loss objective.
 High resolution depth prediction is, however, crucial for safe robot navigation, in particular for autonomous driving where high resolution enables robust long-term perception, prediction, and planning, especially at higher speeds. 
 Furthermore, simply operating at higher image resolutions can be shown to improve overall disparity estimation accuracy (Section~\ref{sec:experiments}).
 We utilize this intuition and propose a deep architecture leveraging super-resolution techniques to improve monocular disparity estimation. 
 


 \begin{figure}[!t]
 \centering
  \includegraphics[trim={0.5cm 0 0 0},clip,width=\columnwidth]{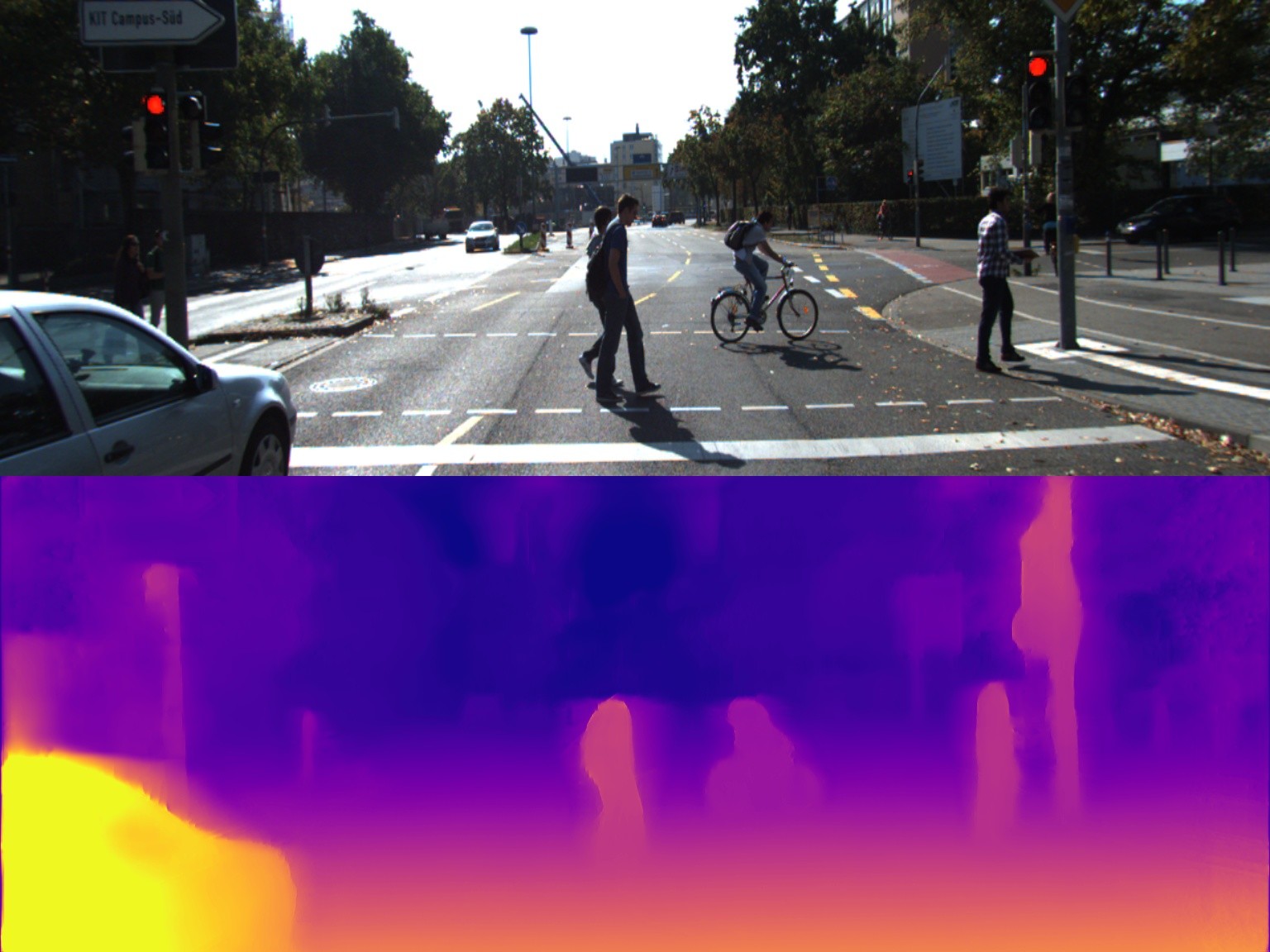} 
 \\
  \includegraphics[trim={0 0 0 21cm},clip,width=0.49\columnwidth]{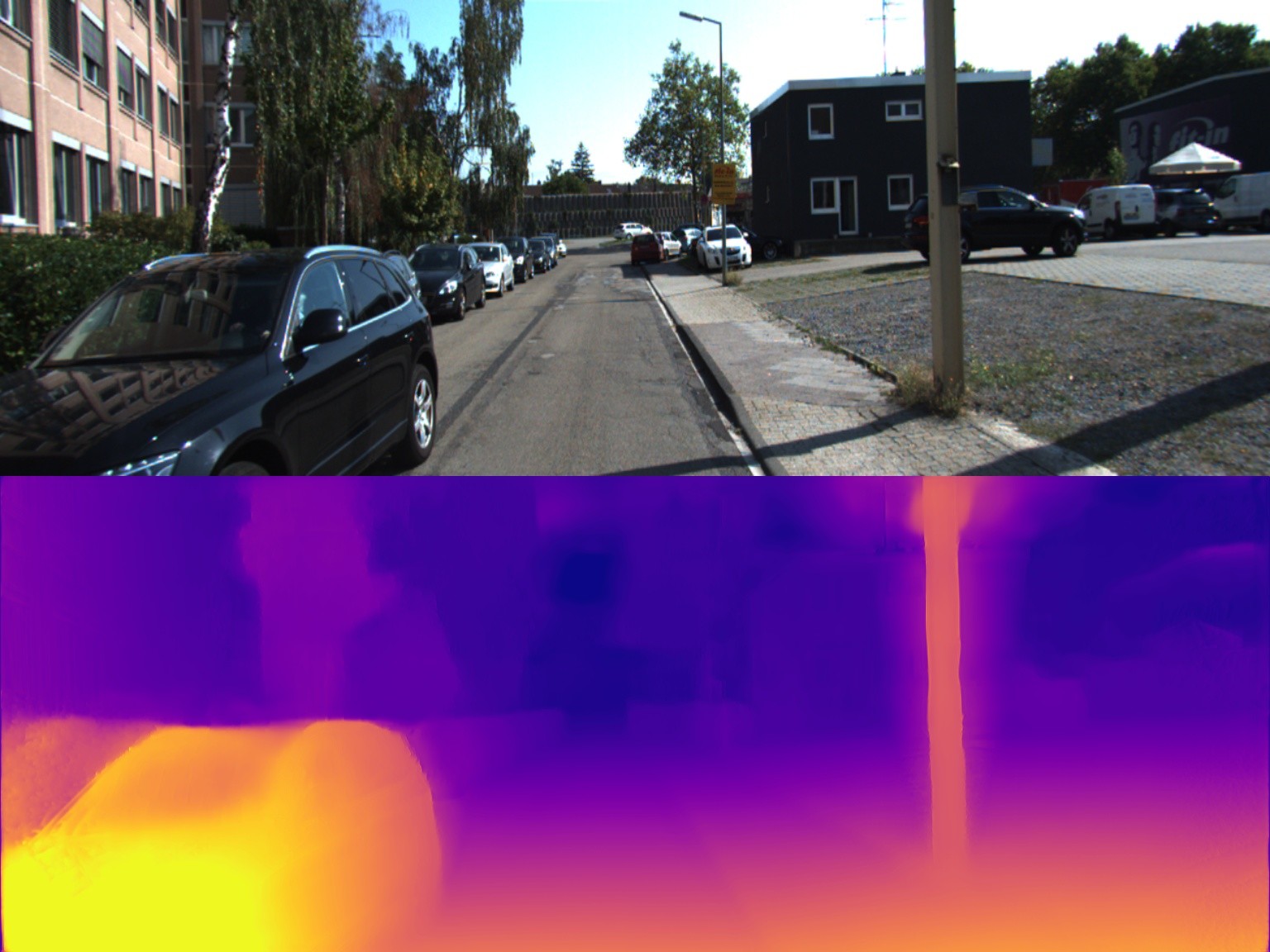} 
  \includegraphics[trim={0 0 0 21cm},clip,width=0.49\columnwidth]{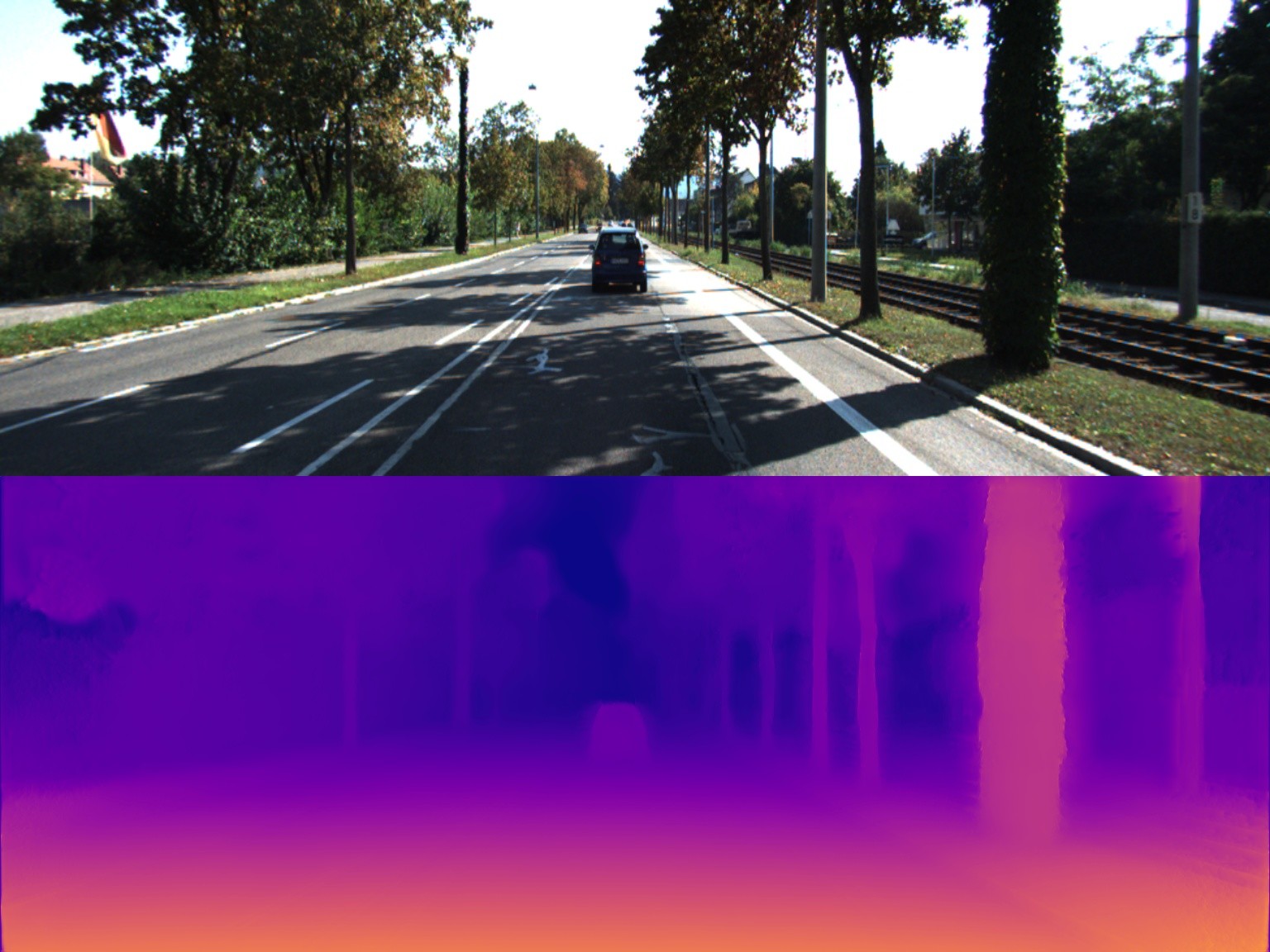}   
  \\
  \includegraphics[width=0.49\columnwidth]{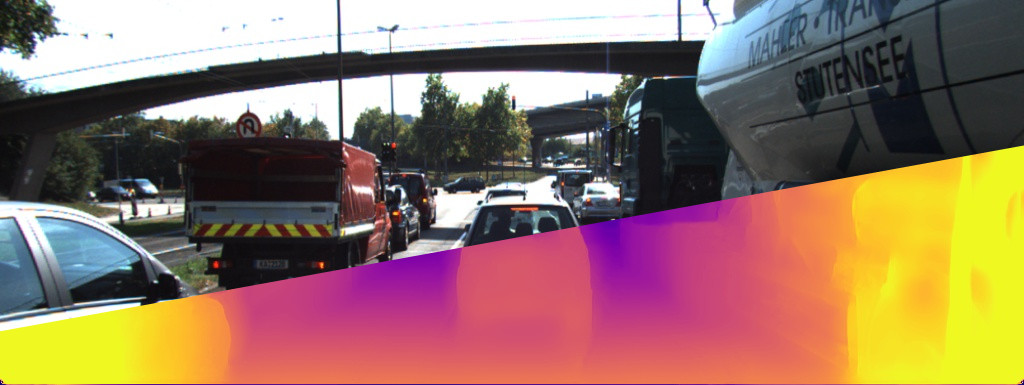}
  \includegraphics[width=0.49\columnwidth]{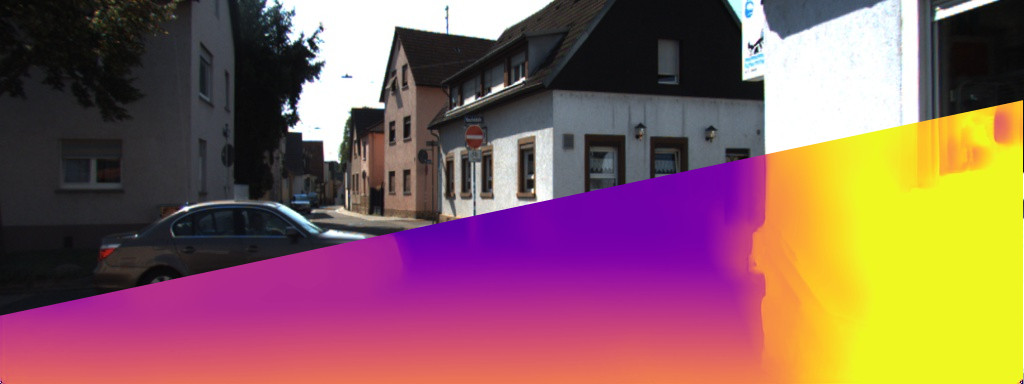}
  \\  
  \includegraphics[width=0.49\columnwidth]{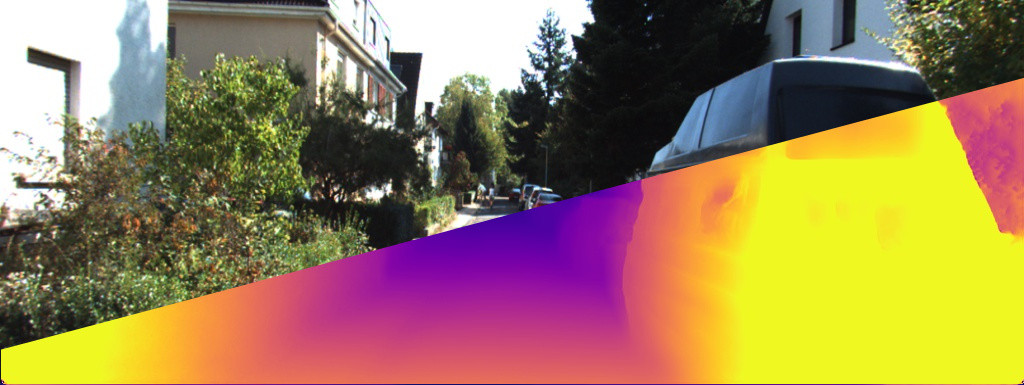}
  \includegraphics[width=0.49\columnwidth]{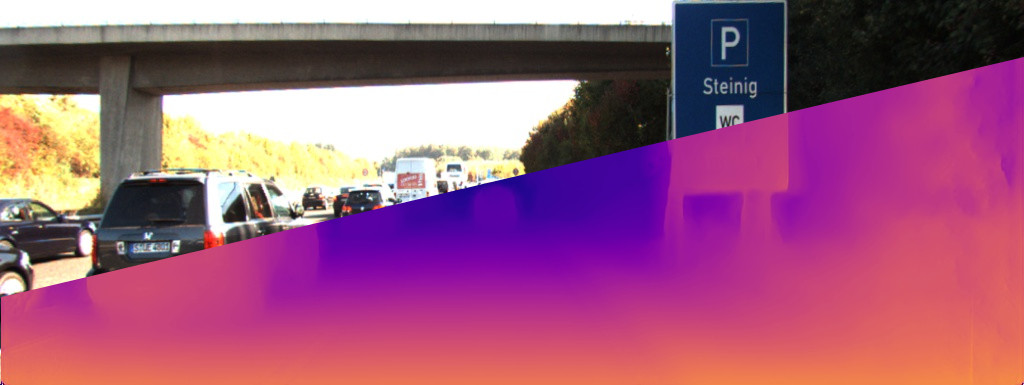}
  \\  
  \caption{Illustration of the accurate and crisp disparities produced by our method from a single monocular image. Our approach combines techniques from Single-Image Super-Resolution (SISR)~\cite{shi2016real} and spatial transformer networks (STN)~\cite{jaderberg2015spatial} to estimate high-resolution, and accurate super-resolved disparity maps.}
  \label{fig:intro-fig}
  \end{figure}

\textbf{Contributions:}
We propose to use \textit{subpixel-convolutional layers} to effectively and accurately super-resolve disparities from their lower-resolution outputs, thereby replacing the deconvolution or resize-convolution~\cite{odena2016deconvolution} up-sampling layers typically used in the disparity decoder networks~\cite{monodepth17,zhou2017unsupervised}.
Second, we introduce a \textit{differentiable flip-augmentation layer} that allows the disparity model to learn an improved prior for disparities at image boundaries in an end-to-end fashion. This results in improved test-time depth predictions with reduced artifacts and occluded regions, effectively removing the need for additional post-processing steps typically used in other methods~\cite{godard2017unsupervised,yin2018geonet}. 
We train our monocular disparity estimation network in a \textit{self-supervised} manner using a synchronized stream of stereo imagery, relieving the need for ground truth depth labels. 
We show that our proposed layers provide significant performance gains to the overall monocular disparity estimation accuracy (Figure~\ref{fig:intro-fig}), especially at higher image resolutions as we detail in our experiments on the public KITTI benchmark.

\section{Related Work}
\label{sec:related-work}

The problem of depth estimation from a single RGB image is an ill-posed inverse problem. Many 3D scenes can indeed correspond to the same 2D image, for instance because of scale ambiguities.
Therefore, solving this problem requires the use of strong priors, in the form of geometric~\cite{godard2017unsupervised,zhou2017unsupervised,yang2018deep}, ordinal~\cite{fu2018deep}, or temporal constraints~\cite{godard2018digging,zhou2017unsupervised,mahjourian2018unsupervised}.
Another effective form of strong prior knowledge is statistical in nature: powerful representations learned by deep neural networks trained on large scale data. CNNs have indeed shown consistent progress towards robust scene depth and 3D reconstruction~\cite{yang2018deep,bloesch2018codeslam,zhoudeeptam2018}. State-of-the-art approaches in leveraging both data and structural constraints mostly differ by the type of data and supervision used.


\textbf{Supervised Depth Estimation}~
Saxena et al.~\cite{saxena2009make3d} proposed one of the first monocular depth estimation techniques, learning patch-based regression and relative depth interactions using Markov Random Fields trained on ground truth laser scans.
Eigen et al.~\cite{eigen2014depth} proposed a multiscale CNN architecture trained on ground truth depth maps by minimizing a scale-invariant loss.
Fully supervised deep learning-based approaches have since  then continuously advanced the state of the art through various architecture and loss improvements~\cite{mayer2016large,zbontar2015computing,ummenhofer2017demon,kendall2017end}.
Semi-supervised methods~\cite{kuznietsov2017semi,yang2018deep} can, in theory, alleviate part of the labeling cost. However, so far they have only been evaluated when using similar amounts of labeled data, reporting significant improvements nonetheless.
Another alternative to circumvent the difficulty of getting ground truth depth maps consists of using synthetic data coming from a simulator~\cite{mayer2018what}, trading off the labeling problem for a domain adaptation and virtual scene creation one.

\textbf{Self-supervised Depth Estimation}~
Procuring large amounts of ground truth depth or disparity maps is expensive, often requiring an entirely different data collection platform than the target robotic deployment platform.
\textit{Self-supervised} learning methods have recently proven to be a promising direction to circumvent this major limitation.
Recent advancements, for instance Spatial Transformer Networks~\cite{jaderberg2015spatial}, have indeed opened the door to a variety of differentiable geometric constraints used as learning objectives capturing key scene properties characterizing optical flow~\cite{yin2018geonet,meister2017unflow}, depth~\cite{godard2017unsupervised,garg2016unsupervised,li2017undeepvo,mahjourian2018unsupervised}, and camera pose~\cite{zhou2017unsupervised,li2017undeepvo}.
Self-supervised approaches thus typically focus on engineering the learning objective, for instance by treating view-synthesis as a proxy task~\cite{zhou2016view,flynn2016deepstereo,godard2017unsupervised,godard2018digging,li2017undeepvo,fei2018geo}.
%
%
Related works also typically explore different architectures, for instance using shared encoders~\cite{godard2018digging} for simultaneous depth and pose estimation.
In contrast, our contributions rely on changing fundamental building blocks of the depth prediction CNN architecture using ideas developed initially for super-resolution~\cite{shi2016real}, or transforming post-processing heuristics into trainable parts of our model.

\newpage
\section{Self-supervised, Super-Resolved Monocular Depth Estimation}
\label{sec:procedure}

The goal of monocular depth estimation is the recovery of a function $f_z: I \to D_z$, that predicts the depth $\hat{z} = f_z(I(p))$ for every pixel $p$ in the given input image $I$. In this work, we learn to recover the disparity estimation function $f_d: I \to D$ in a \textit{self-supervised} manner from a synchronized stereo camera (Section~\ref{sec:proc-monodepth-architecture}). Given $f_d$, we can  estimate the disparity $\hat{d} = f_d(I(p))$ for every pixel $p$ in the input image $I$, with the metric depth $\hat{z}$ estimated via $\hat{z} = \frac{fB}{\hat{d}}$. Both the camera focal length $f$ and stereo baseline $B$ are assumed to be known while training. 
\begin{figure*}[!t]
  \centering
  {
   {\renewcommand{\arraystretch}{0.2}
    \setlength{\tabcolsep}{0.2mm}
    \begin{tabular}{cccccc}
    Input image & Ours & MonoDepth \cite{monodepth17} & GeoNet \cite{yin2018geonet} & SfMLearner \cite{zhou2017unsupervised} & Vid2Depth \cite{mahjourian2018unsupervised} \\
    \includegraphics[width=0.34\columnwidth,height=1.1cm]{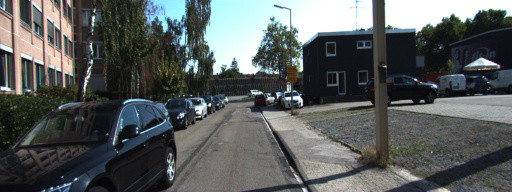}&
    \includegraphics[width=0.34\columnwidth,height=1.1cm]{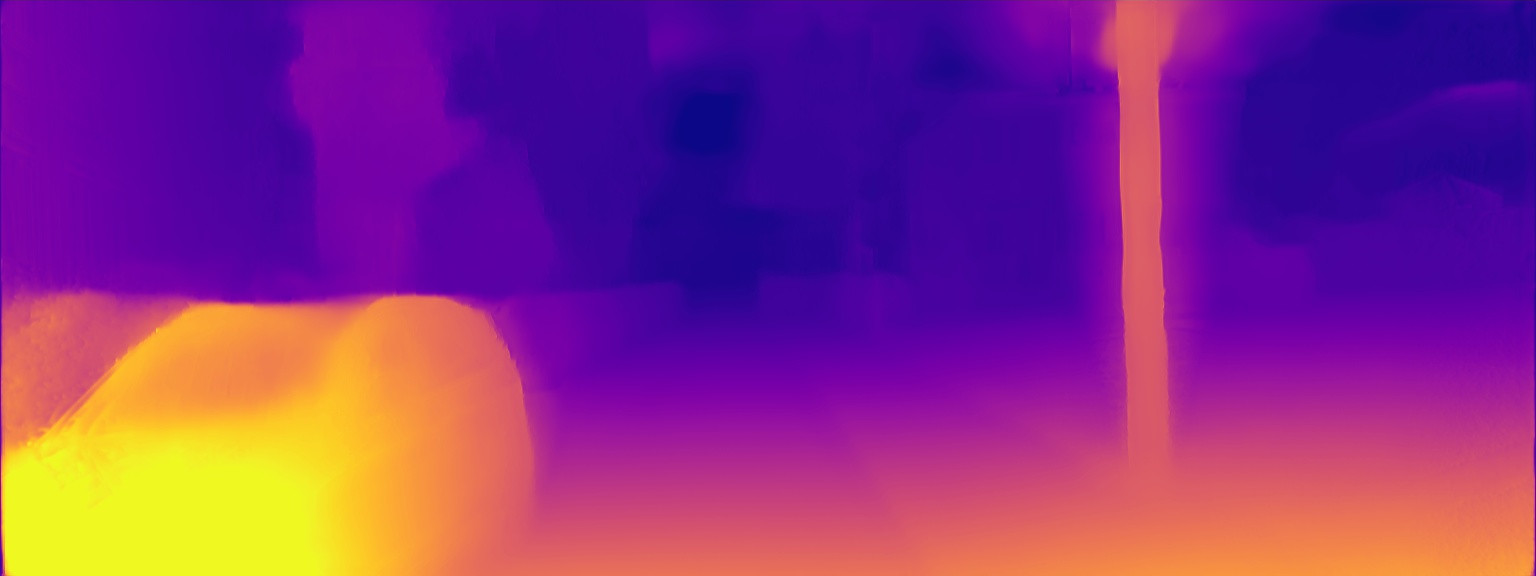}&
    \includegraphics[width=0.34\columnwidth,height=1.1cm]{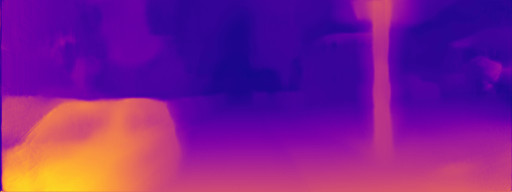}&
    \includegraphics[width=0.34\columnwidth,height=1.1cm]{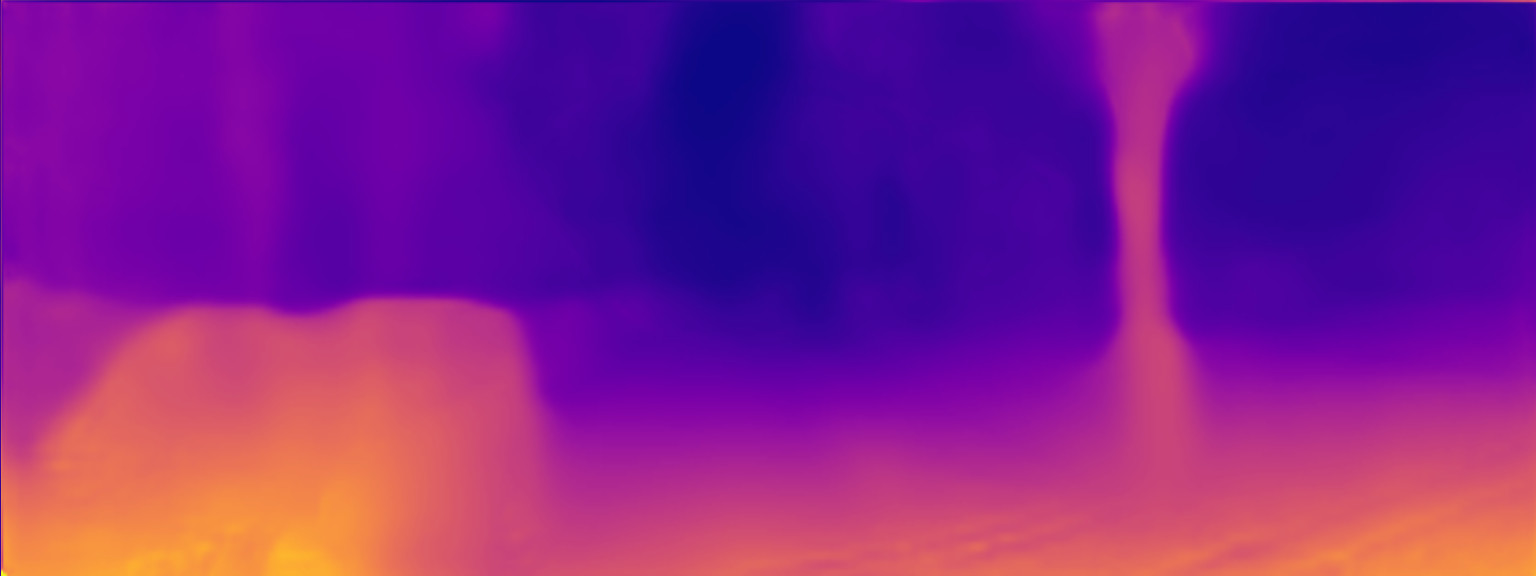}&
    \includegraphics[width=0.34\columnwidth,height=1.1cm]{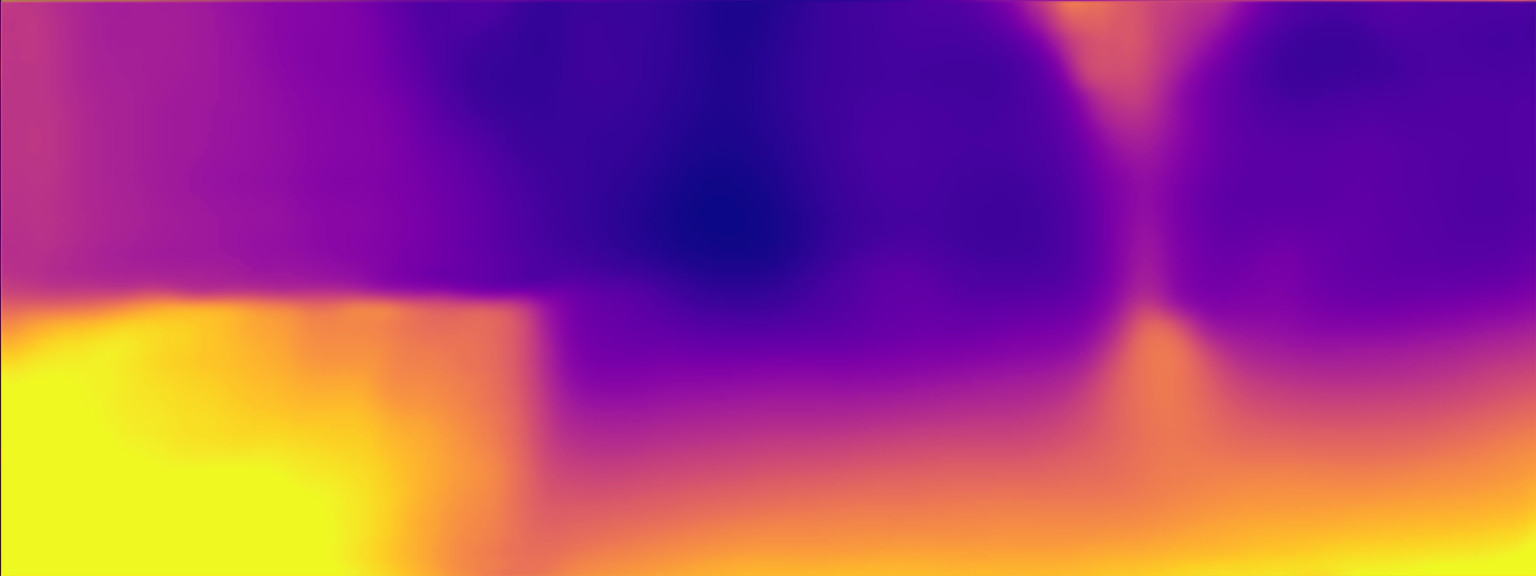}&
    \includegraphics[width=0.34\columnwidth,height=1.1cm]{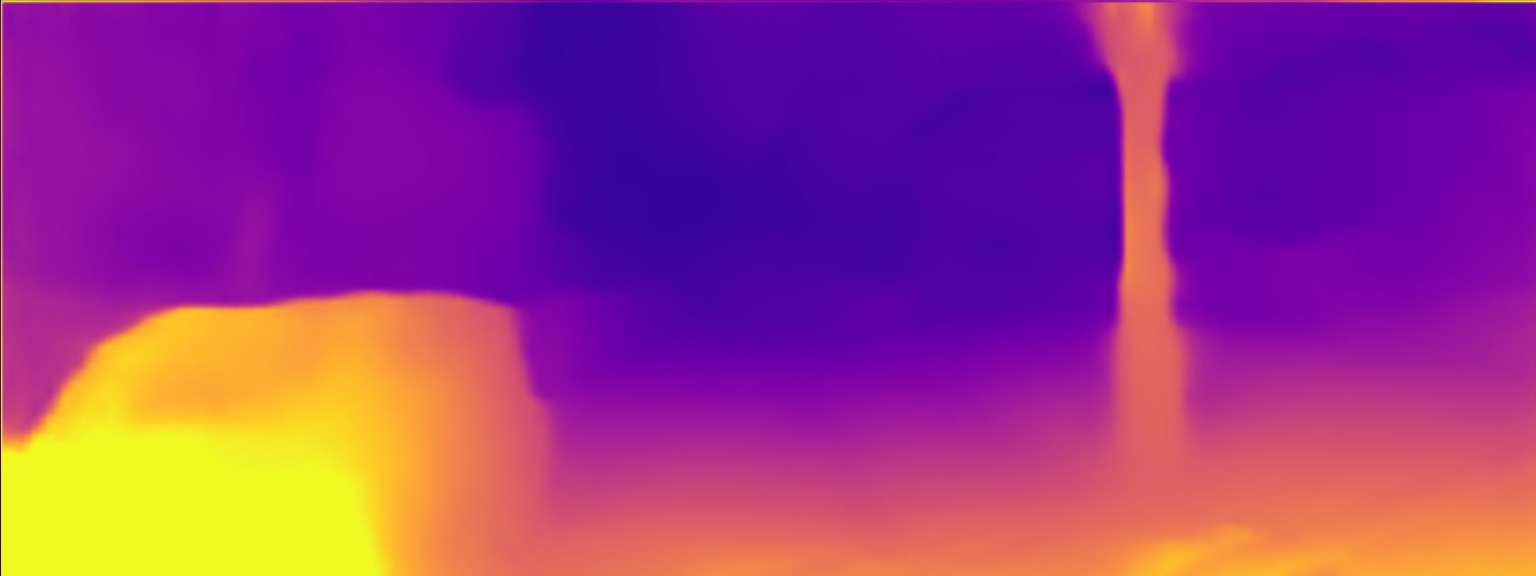}\\
    \includegraphics[width=0.34\columnwidth,height=1.1cm]{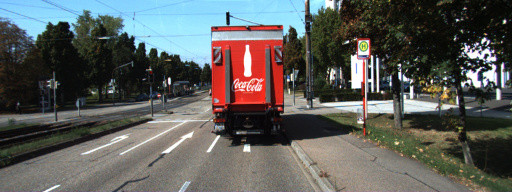}&
    \includegraphics[width=0.34\columnwidth,height=1.1cm]{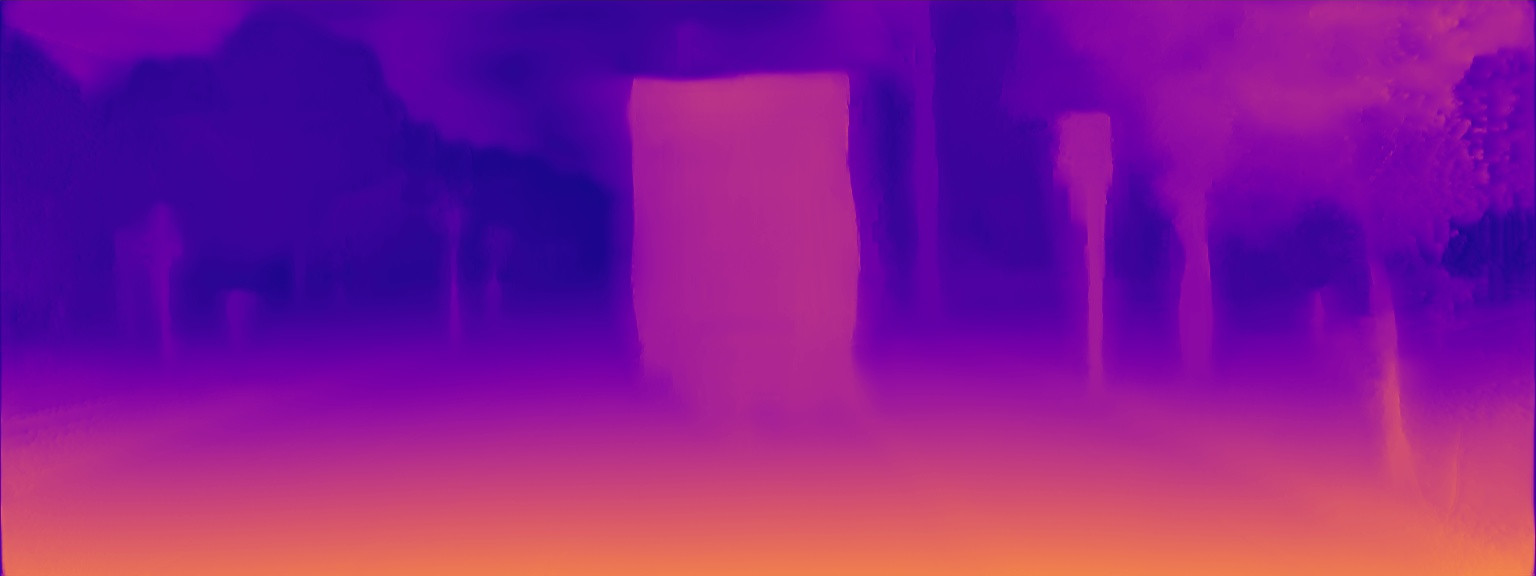}&
    \includegraphics[width=0.34\columnwidth,height=1.1cm]{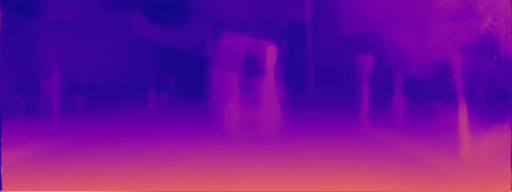}&
    \includegraphics[width=0.34\columnwidth,height=1.1cm]{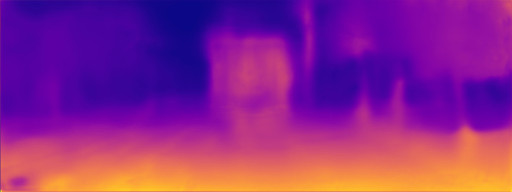}&
    \includegraphics[width=0.34\columnwidth,height=1.1cm]{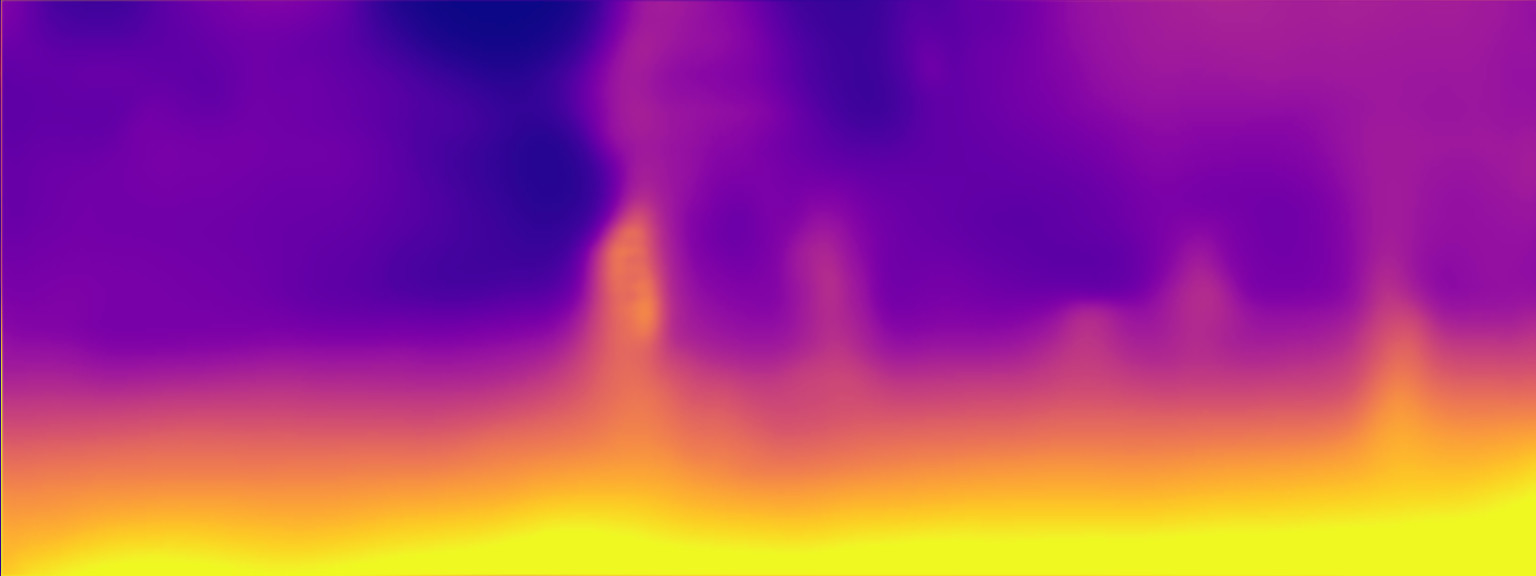}&
    \includegraphics[width=0.34\columnwidth,height=1.1cm]{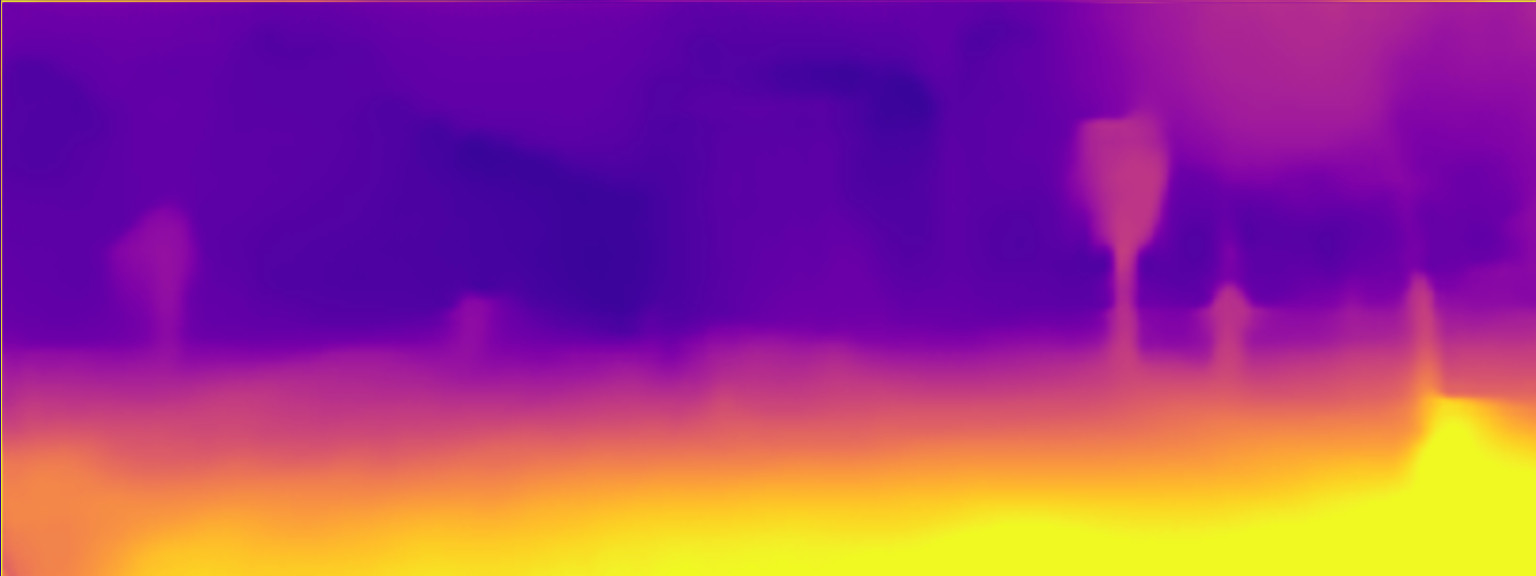}\\
    \includegraphics[width=0.34\columnwidth,height=1.1cm]{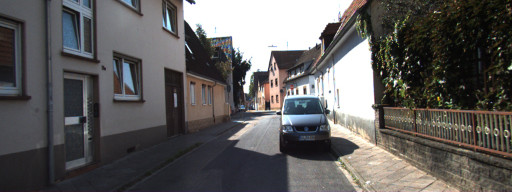}&
    \includegraphics[width=0.34\columnwidth,height=1.1cm]{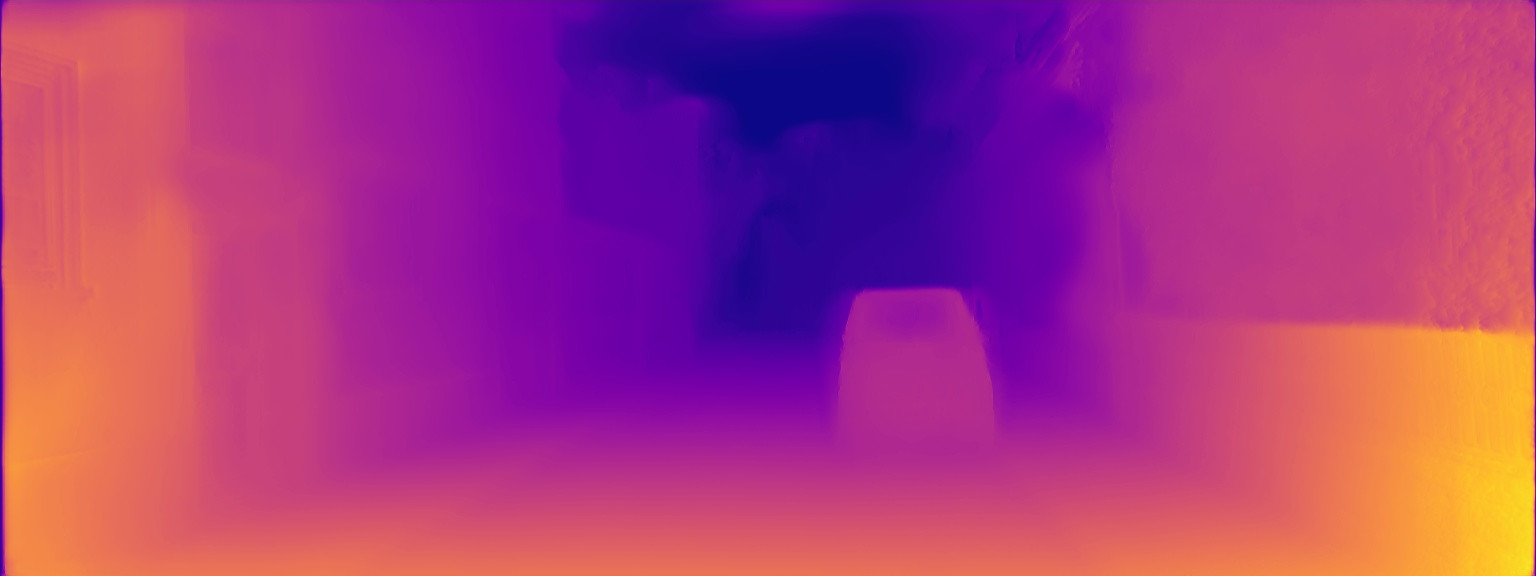}&
    \includegraphics[width=0.34\columnwidth,height=1.1cm]{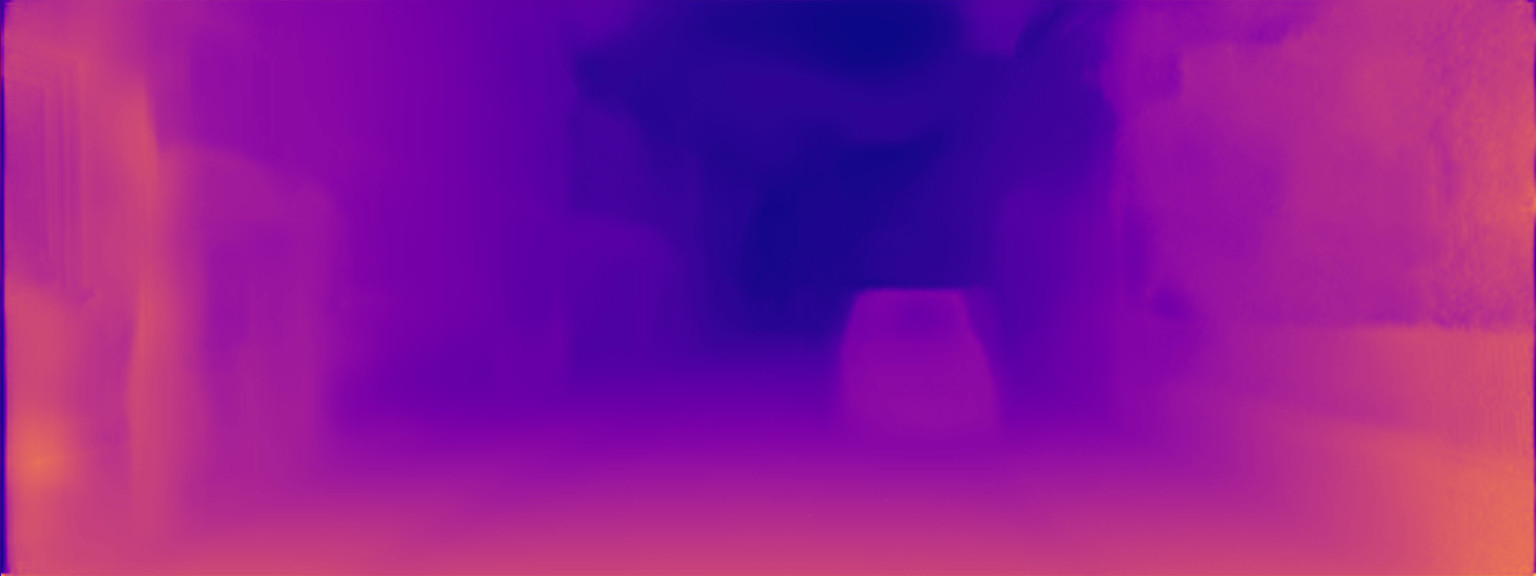}&
    \includegraphics[width=0.34\columnwidth,height=1.1cm]{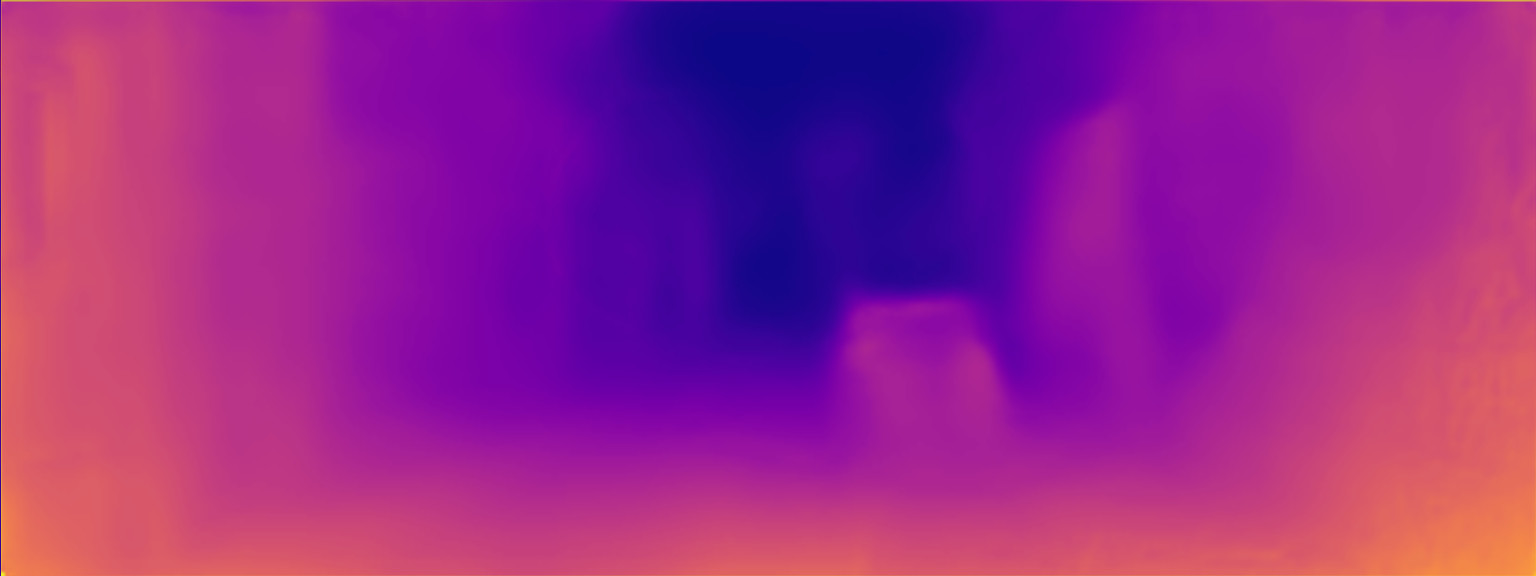}&
    \includegraphics[width=0.34\columnwidth,height=1.1cm]{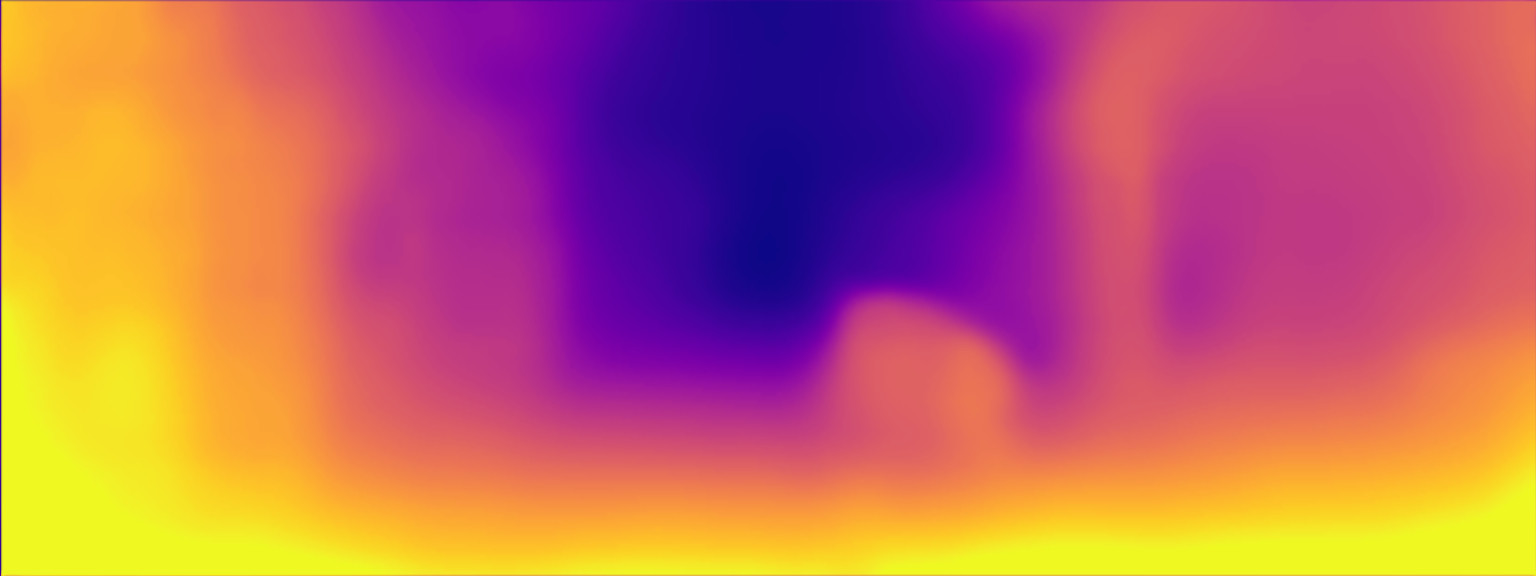}&
    \includegraphics[width=0.34\columnwidth,height=1.1cm]{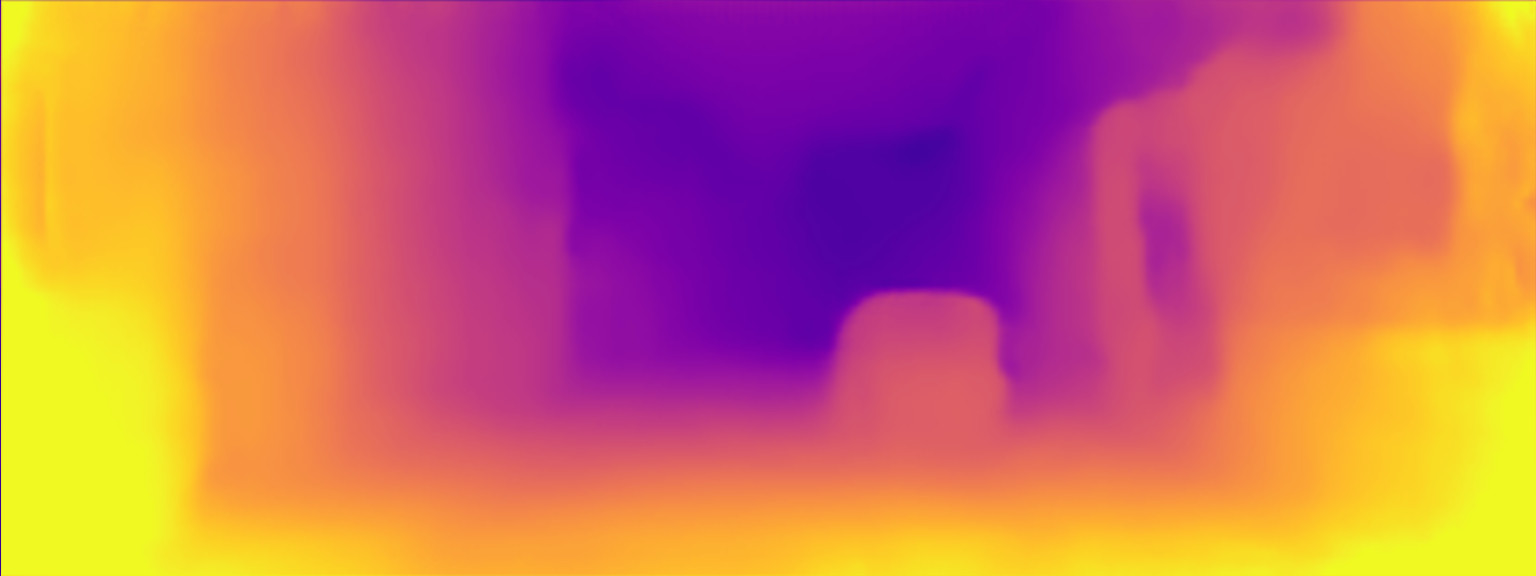}\\
    \includegraphics[width=0.34\columnwidth,height=1.1cm]{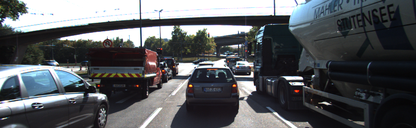}&
    \includegraphics[width=0.34\columnwidth,height=1.1cm]{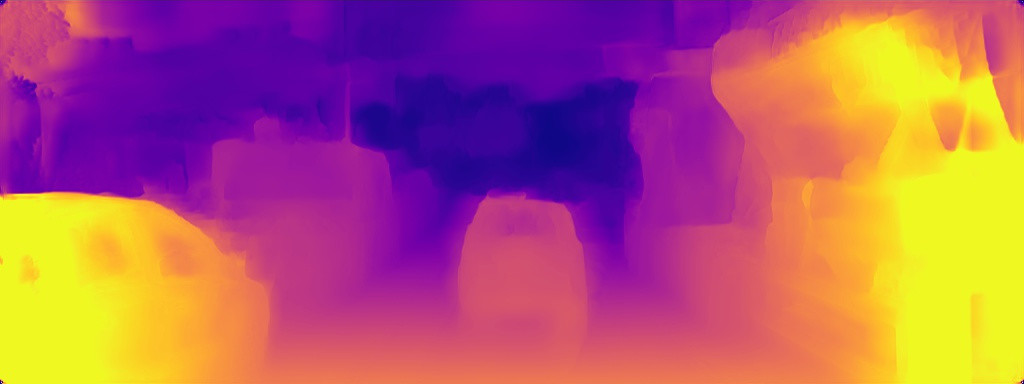}&
    \includegraphics[width=0.34\columnwidth,height=1.1cm]{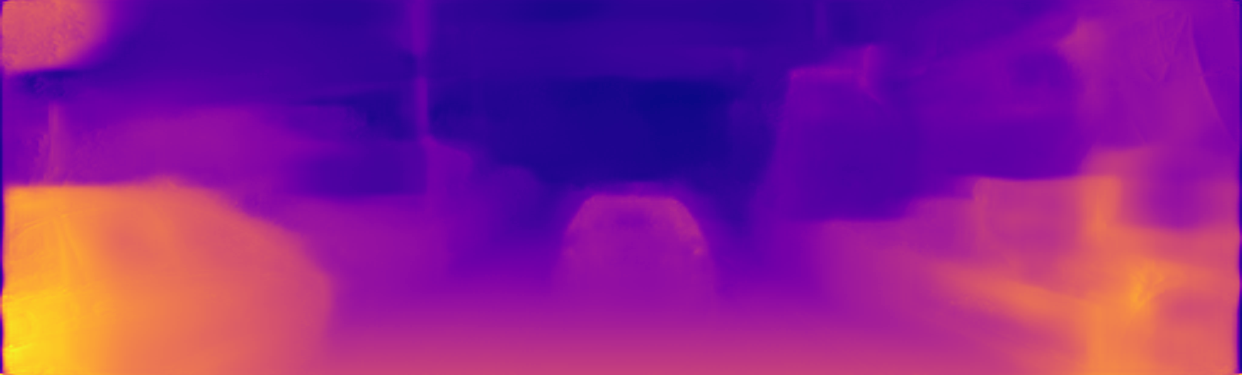}&
    \includegraphics[width=0.34\columnwidth,height=1.1cm]{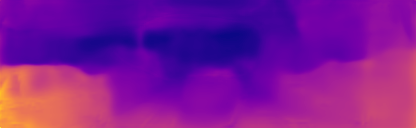}&
    \includegraphics[width=0.34\columnwidth,height=1.1cm]{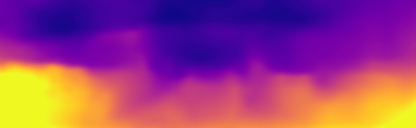}&
    \includegraphics[width=0.34\columnwidth,height=1.1cm]{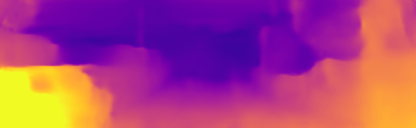}\\
    \includegraphics[width=0.34\columnwidth,height=1.1cm]{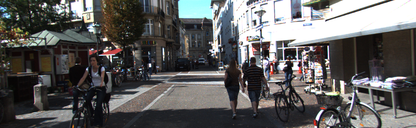}&
    \includegraphics[width=0.34\columnwidth,height=1.1cm]{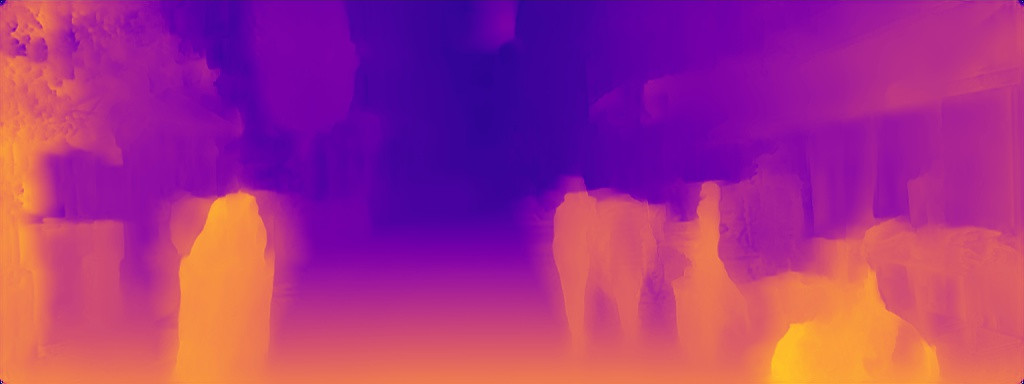}&
    \includegraphics[width=0.34\columnwidth,height=1.1cm]{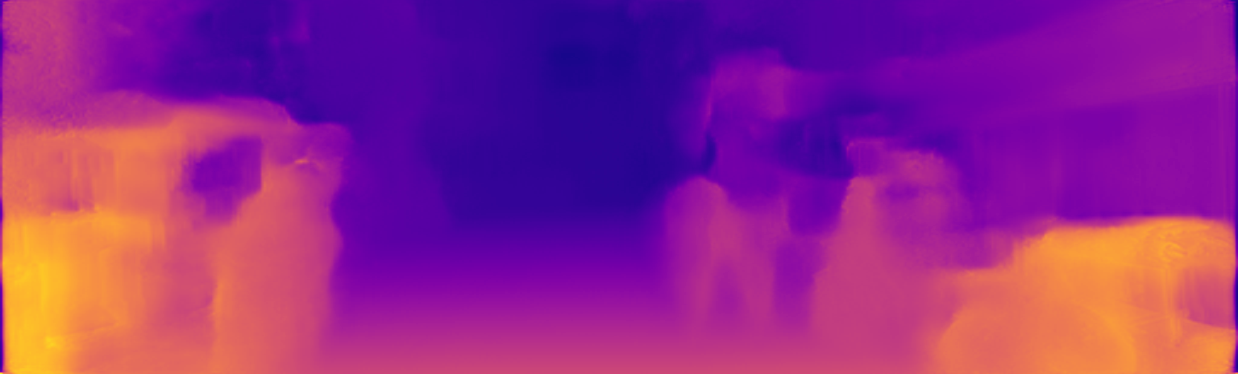}&
    \includegraphics[width=0.34\columnwidth,height=1.1cm]{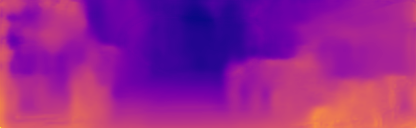}&
    \includegraphics[width=0.34\columnwidth,height=1.1cm]{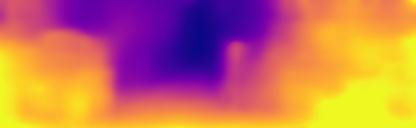}&
    \includegraphics[width=0.34\columnwidth,height=1.1cm]{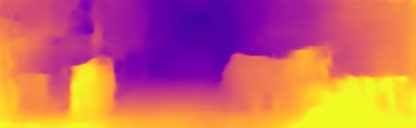}\\
    \includegraphics[width=0.34\columnwidth,height=1.1cm]{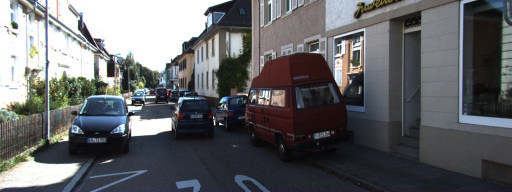}&
    \includegraphics[width=0.34\columnwidth,height=1.1cm]{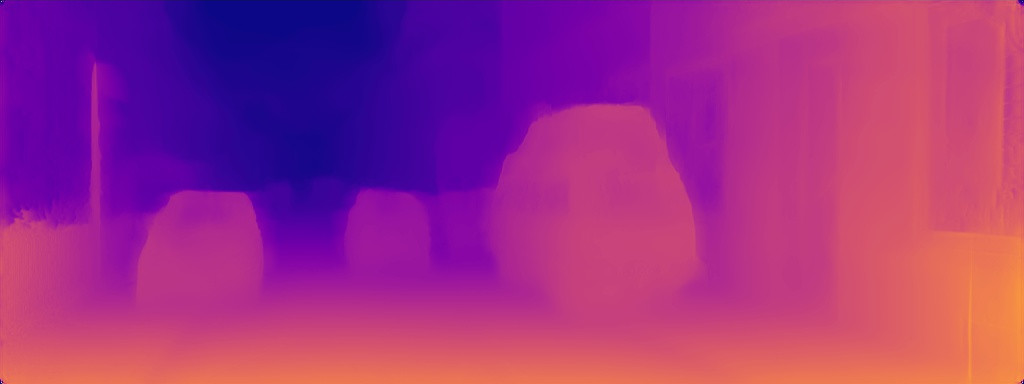}&
    \includegraphics[width=0.34\columnwidth,height=1.1cm]{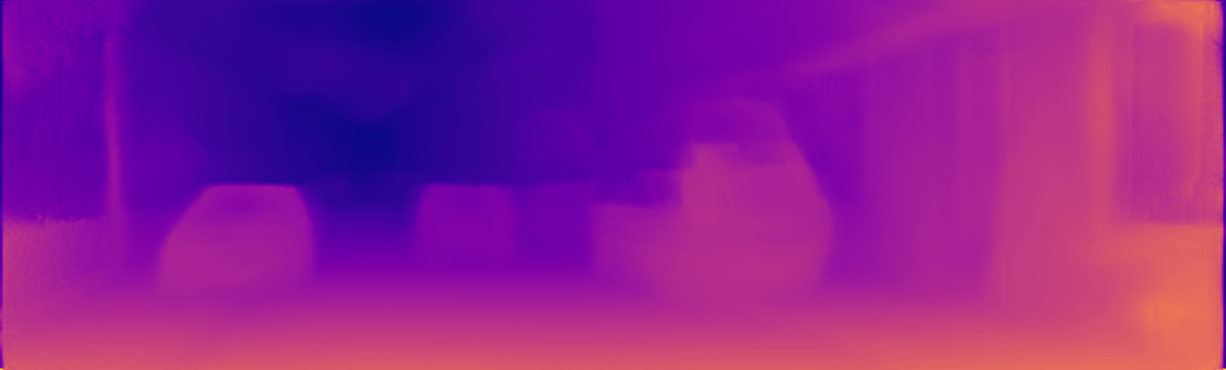}&
    \includegraphics[width=0.34\columnwidth,height=1.1cm]{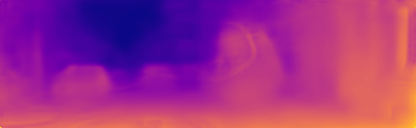}&
    \includegraphics[width=0.34\columnwidth,height=1.1cm]{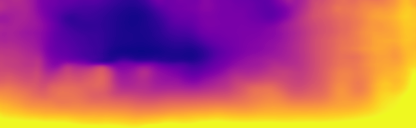}&
    \includegraphics[width=0.34\columnwidth,height=1.1cm]{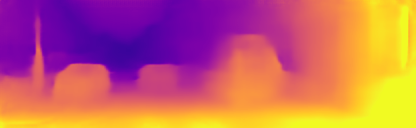}\\
    \includegraphics[width=0.34\columnwidth,height=1.1cm]{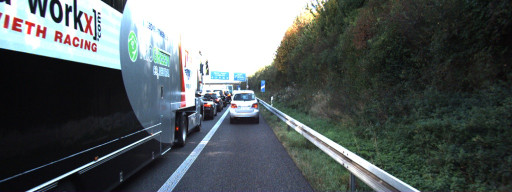}&
    \includegraphics[width=0.34\columnwidth,height=1.1cm]{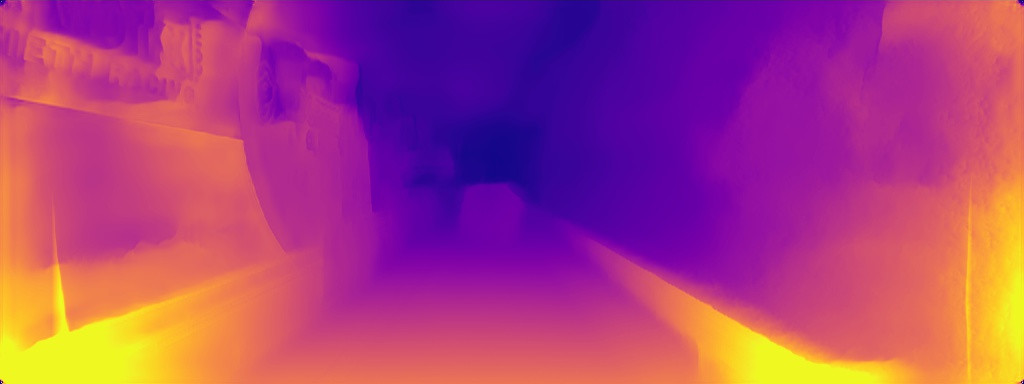}&
    \includegraphics[width=0.34\columnwidth,height=1.1cm]{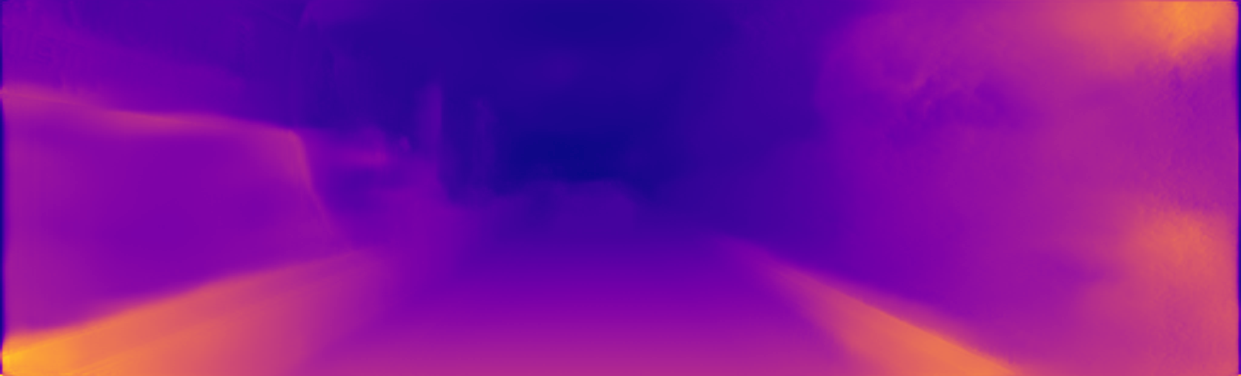}&
    \includegraphics[width=0.34\columnwidth,height=1.1cm]{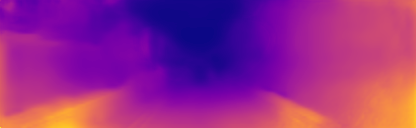}&
    \includegraphics[width=0.34\columnwidth,height=1.1cm]{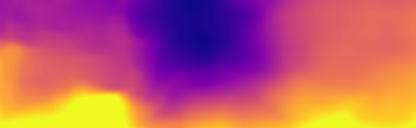}&
    \includegraphics[width=0.34\columnwidth,height=1.1cm]{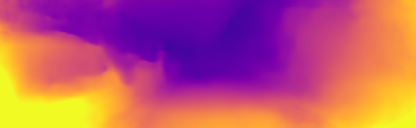}\\
    \includegraphics[width=0.34\columnwidth,height=1.1cm]{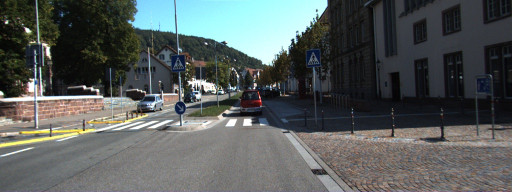}&
    \includegraphics[width=0.34\columnwidth,height=1.1cm]{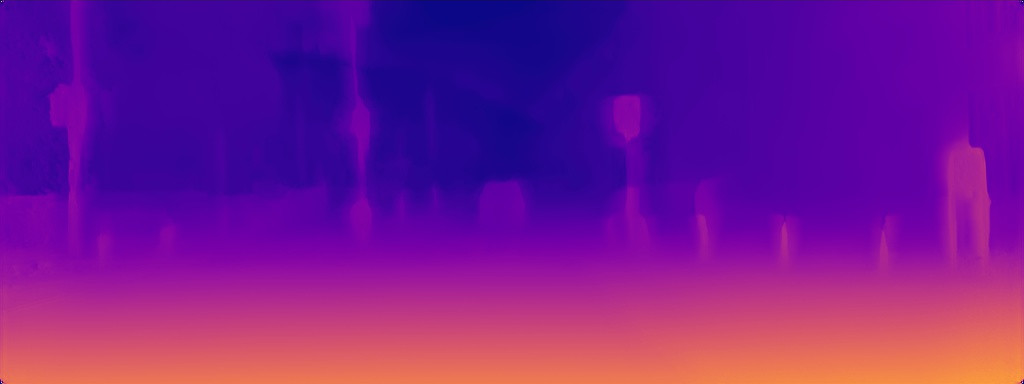}&
    \includegraphics[width=0.34\columnwidth,height=1.1cm]{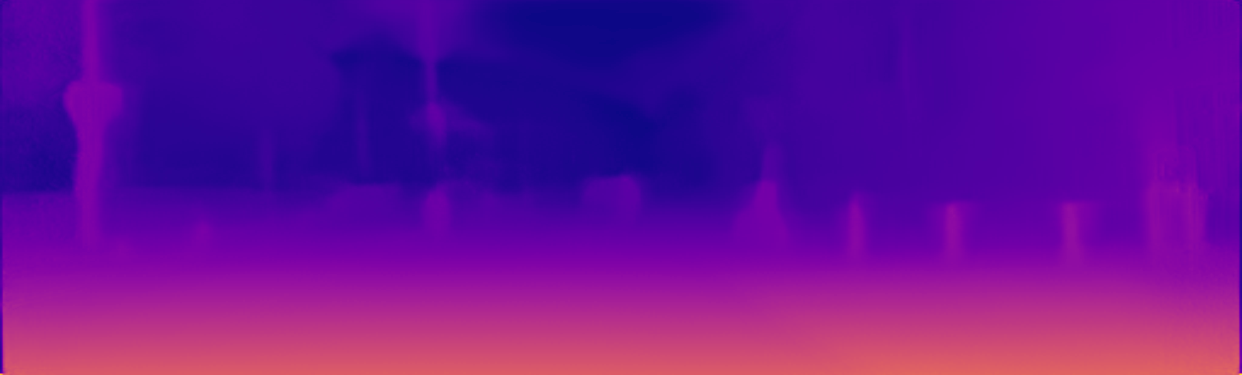}&
    \includegraphics[width=0.34\columnwidth,height=1.1cm]{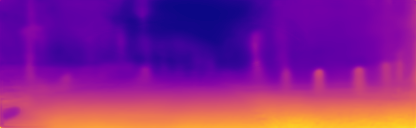}&
    \includegraphics[width=0.34\columnwidth,height=1.1cm]{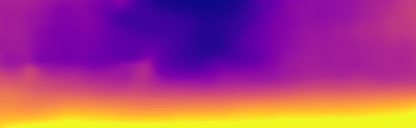}&
    \includegraphics[width=0.34\columnwidth,height=1.1cm]{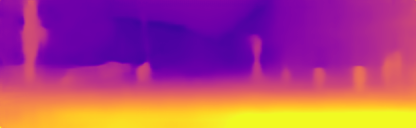}\\
\end{tabular}}}
  \caption{Illustrated above are qualitative comparisons of our proposed self-supervised, super-resolved monocular depth estimation method with previous state-of-the-art methods. We show that our approach produces qualitatively better depth estimates with crisp boundaries. Our method also correctly reconstructs thin and far-off objects reliably compared to previous methods that tend to only estimate shadow artifacts in such regions. }
  \label{fig:disparity-illustration2}
\vspace{-6mm}
\end{figure*}

\subsection{Monocular Depth Network Architecture}
\label{sec:proc-monodepth-architecture}
Our disparity estimation model builds upon the popular DispNet~\cite{mayer2016large} architecture. Following Godard et al.~\cite{godard2017unsupervised}, we make similar modifications to the encoder-decoder network with skip connections~\cite{long2015fully} between the encoder's activation blocks. However, unlike the left-right (LR) disparity architecture~\cite{godard2017unsupervised}, the model outputs a single disparity channel. We further extend this base architecture to incorporate two key components detailed in the following sections. 

\subsubsection{Sub-pixel Convolution for Depth Super-Resolution}
Recent methods that employ multi-scale disparity estimation utilize deconvolutions, resize-convolutions~\cite{odena2016deconvolution} or naive interpolation operators (for e.g. bilinear, nearest-neighbor) to up-sample the lower-resolution disparities to their target image resolution.
However these methods perform the interpolation in the high-resolution space, and are limited in their representational capacity for disparity super-resolution. Inspired by recent CNN-based methods for Single-Image-Super-Resolution (SISR)~\cite{shi2016real}, we introduce a sub-pixel convolutional layer based on ESPCN~\cite{shi2016real} for depth super-resolution that accurately synthesizes the high-resolution disparities from their corresponding low-resolution multi-scale model outputs. 
This effectively replaces the disparity interpolation layer, while learning relevant low-resolution convolutional features that can perform high-quality disparity synthesis. We swap the resize-convolution branches from each of the 4 pyramid scales in the disparity network with the sub-pixel convolutional branch consisting of a sequence of 4 consecutive 2D convolutional layers with 32, 32, 32, 16 layers with 1 pixel stride, each followed by ReLu activations. The final convoluational output is then re-mapped to the target depth resolution via a pixel re-arrange operation, resulting in an efficient sub-pixel convolutional operation as described in~\cite{shi2016real}.  


\subsubsection{Differentiable Flip Augmentation}
In stereopsis, due to the lack of observable left image boundary scene points in the right image, the disparity model will inevitably learn a poor prior on boundary-pixels. To circumvent this behavior, previous methods~\cite{godard2017unsupervised,godard2018digging} include a post-processing step that alpha-blends the disparity images from the image and its horizontally flipped version. While this significantly reduces visual artifacts around the image boundary and improves overall accuracy, it however decouples the final disparity estimation from the training. To this end, we replace this step with a \textit{differentiable flip-augmentation layer} within the disparity estimation model itself, allowing us to fine-tune disparities in an end-to-end fashion. By leveraging the differentiable image-rendering in~\cite{jaderberg2015spatial} to revert the flipped disparity, the model performs the forward pass with the identical model on both the original and horizontally flipped images. The outputs are fused together in a differentiable manner with a pixel-wise mean operation while handling the borders similar to~\cite{godard2017unsupervised}. 



\subsection{Self-supervising Depth with Stereopsis}  
\label{sec:proc-ssl-depth}
Following~\cite{godard2017unsupervised,garg2016unsupervised,zhou2017unsupervised}, we formulate the disparity estimation as a photometric error minimization problem across multiple camera views. We define $D_t$ as the disparity image for the corresponding target image $I_t$, and re-cast the disparity estimation implicitly as an image synthesis task of a new source image $I_s$. The photometric error is then re-written as the minimization of pixel-intensity difference between the target image $I_t$, and the synthesized target image re-projected from the source image's view $\hat{I_t} = I_s(p_s)$~\cite{jaderberg2015spatial}. Here, $p_s \sim K \mathbf{x}_{t \to s}  \hat{D_t}(p_t) K^{-1} p_t$ is the source pixel derived from re-projecting the target pixel $p_t$ in the source image's view $\mathbf{x}_s$, with $\mathbf{x}_{t \to s}$ describing the relative transformation between the target image view pose $\mathbf{x}_t$ and source image view pose $\mathbf{x}_s$. The disparity estimation model $f_d$ parametrized by $\theta_d$ is defined as: 
\begin{align}
  \hat{\theta_D} = \argmin_{\theta_D} \sum_{s \in S} \mathcal{L}_{D}(I_t, \hat{I_t}; \theta_D)
  \label{eq:disp-loss-minimization}
\end{align}
where $s \in S$ are all the disparate views available for synthesizing the target image $I^t$. In the case of stereo cameras, $\mathbf{x}_{s \to t}$ in Equation~\ref{eq:disp-loss-minimization} is known a-priori, and directly incorporated as a constant within the overall minimization objective. The overall loss $\mathcal{L}_d$ comprises of 3 terms: 
\begin{align}
  \mathcal{L}_{D}(I_t,\hat{I_t}) = \mathcal{L}_p(I_t,\hat{I_t}) + \lambda_1~\mathcal{L}_s(I_t) + \lambda_2~\mathcal{L}_o(I_t)
  \label{eq:overall-loss-disp}
\end{align}


\textbf{Appearance Matching Loss}~~Following~\cite{godard2017unsupervised}, the pixel-level similarity between the target image $I_t$ and the synthesized target image $\hat{I_t}$ is estimated using the Structural Similarity (SSIM)~\cite{wang2004image} term combined with an L1 photometric-term, inducing an overall loss given by Equation~\ref{eq:loss-photo} below.   
\begin{align}
  \mathcal{L}_{p}(I_t,\hat{I_t}) = \alpha_1~\frac{1 - \text{SSIM}(I_t,\hat{I_t})}{2} + (1-\alpha_1)~\| I_t - \hat{I_t} \|
  \label{eq:loss-photo}
\end{align}

\textbf{Disparity Smoothness Loss}~~In order to regularize the disparities in textureless low-image gradient regions, we incorporate an edge-aware term (Equation ~\ref{eq:loss-disp-smoothness}), similar to~\cite{godard2017unsupervised,yang2018deep,li2017undeepvo}. The effect of each of the pyramid-levels is decayed by a factor of 2 on downsampling, starting with a weight of 1 for the $0^\text{th}$ pyramid level. 
\begin{align}
  \mathcal{L}_{s}(I_t) = | \delta_x d_t | e^{-|\delta_x I_t|} + | \delta_y d_t | e^{-|\delta_y I_t|}
  \label{eq:loss-disp-smoothness}
\end{align}

\textbf{Occlusion Regularization Loss}~~We adopt the occlusion regularization term (similar to ~\cite{yang2018deep}) to minimize the shadow areas generated in the disparity map, especially across high gradient disparity regions. By inducing an L1-loss over the disparity estimate, this term encourages background depths (i.e. lower disparities) by penalizing the total sum of absolute disparities in the estimate. 
\begin{align}
  \mathcal{L}_{o}(I_t) = | d_t | 
  \label{eq:loss-occ-reg}
\end{align}
The photometric, disparity smoothness and occlusion regularization losses are combined in a final loss (Equation~\ref{eq:overall-loss-disp}) which is averaged per-pixel, pyramid-scale and image batch during training.

\section{Experiments}
\label{sec:experiments}

\subsection{Dataset}

We use the KITTI~\cite{geiger2013vision} dataset for all our experiments. We compare with previous methods on the standard KITTI Disparity Estimation benchmark. We adopted the training protocols used in Eigen et al.~\cite{eigen2014depth}, and specifically, we used the KITTI \textit{Eigen} splits described in~\cite{eigen2014depth} that contain 22600 training, 888 validation, and 697 test stereo image pairs. We evaluate the disparities estimated using the metrics described in Eigen et al.~\cite{eigen2014depth}. 


 
\subsection{Depth Estimation Performance}
We re-implemented the modified DispNet with skip connections as described in Godard et al.~\cite{godard2017unsupervised} as \textit{our} baseline (Ours), and evaluate it with the proposed sub-pixel convolutional extension (Ours-SP) and the differentiable flip-augmentation (Ours-FA). However, first, we show that by operating in high-resolution regimes, we are able to alleviate the drawbacks of the multi-scale photometric loss function that inadvertently incorporate the losses at extremely low-resolution disparities. 

\subsubsection{Effect of High-Resolution in Disparity Estimation}
\begin{figure}[!h]
\centering
{
\footnotesize
\setlength{\tabcolsep}{0.1em}
\rowcolors{2}{lightgray}{white}
\begin{tabular}{lccccccc}
\toprule
\textbf{Resolution} & 
Abs Rel &
Sq Rel &
RMSE log &
$\delta < 1.25$ &
$\delta < 1.25^2$ &
$\delta < 1.25^3$\vspace{0.5mm}\\
\midrule
~512 x 192  & 0.133 & 1.079 &  0.247 &  0.816 &  0.927 &  0.964\\
1024 x 384 & 0.116 & 0.935 &  0.210 &  0.842 &  0.945 &  \textbf{0.977}\\
1536 x 576 & \textbf{0.114} & \textbf{0.869} &  \textbf{0.209} &  0.849 &  0.945 &  0.976\\
2048 x 768 & 0.116 & 1.055 &  \textbf{0.209} &  \textbf{0.853} &  \textbf{0.948} &  \textbf{0.977}\\
\bottomrule
\end{tabular}\\\vspace{0mm}
}
{
\renewcommand{\arraystretch}{0.4} 
   \begin{tabular}{c}
    \includegraphics[width=0.9\columnwidth]{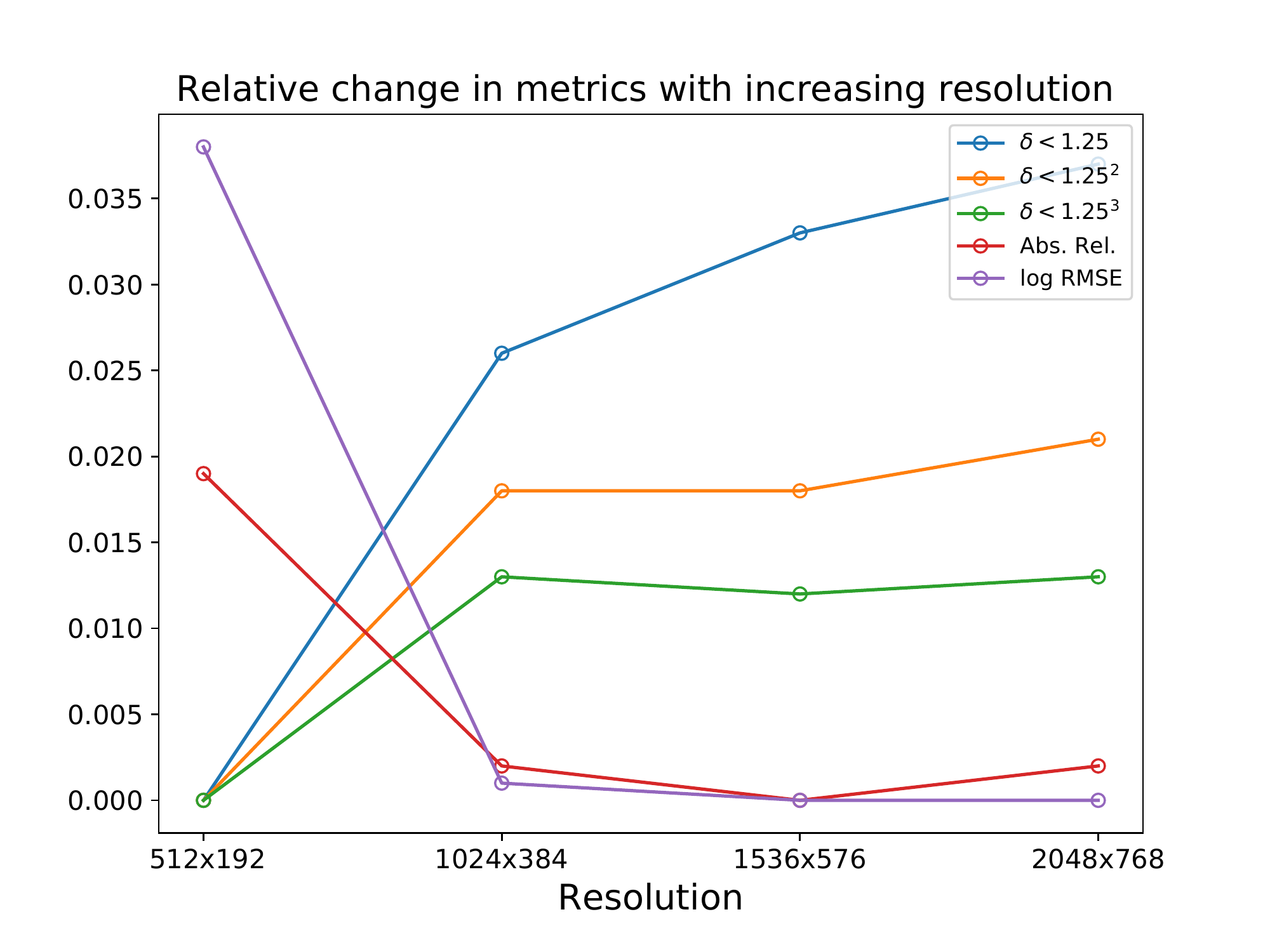}
  \end{tabular}
}
  \caption{\textbf{Effect of high-resolution}: The relative change in the disparity estimation metrics are plotted with increasing input image resolution. We show that by naively increasing the input image resolution, we are able to show considerable improvement (increase in $\delta < z$, and decrease in Abs Rel, Sq Rel, RMSE log metrics) without any changes to the underlying loss function. This motivates us to consider efficient and accurate methods in performing disparity estimation at much higher input image resolutions via sub-pixel convolutions.}
  \label{fig:highres-compare}
\end{figure}

As previously mentioned, the \textit{self-supervised} photometric loss is limited by the image resolution and the corresponding disparities at which they operate. In their recent work~\cite{godard2018digging}, the authors discuss this limitation and up-sample the multi-scaled disparity outputs to their original input image resolution before computing the relevant photometric losses. Using this insight, we first consider estimating disparities at higher-resolutions and use this as our baseline for subsequent experiments. In Figure~\ref{fig:highres-compare}, we show that with increasing input image resolutions of 1024 x 384, 1536 x 576, and 2048 x 768, the disparity estimation performance continues to improve for most metrics including Abs. Rel, Sq Rel. RMSE, and RMSE log. The performance of the baseline approach however saturates at the 1536 x 576 resolution since the original KITTI stereo images are captured at 1392 x 512 pixel resolution. It is however noteworthy that the fraction of the disparities within $\delta < z$ pixels show improvements with even higher input image resolutions indicating that the photometric losses are indeed limited by the disparity resolution. 

\subsubsection{Improving Disparity Estimation with Sub-pixel Convolutions}
Using the insight of operating at high-resolution disparity regimes, we discuss the importance of super-resolving low-resolution disparities estimated within Encoder-Decoder-based disparity networks~\cite{mayer2016large,yang2018deep,ummenhofer2017demon}. With Ours-SP, we are able to achieve a considerable improvement in performance (0.112 abs. rel.) for the same input image resolution over our established baseline Ours (0.116 abs. rel.). Furthermore, we notice that the Sq. Rel., RMSE, $\delta < z$ columns show equally consistent and improved performance over the baseline that utilizes resize-convolutions~\cite{odena2016deconvolution} instead of the proposed sub-pixel convolutional layer for disparity up-sampling. In Table.~\ref{table:accuracy}, we report our disparity estimation results with the proposed \textit{sub-pixel convolutional layer} and the \textit{differentiable flip-augmentation}, illustrating the \textit{state-of-the-art performance} for self-supervised disparity estimation on the KITTI Eigen test set.

\begin{table*}[h]
\centering
{
\small
\setlength{\tabcolsep}{0.3em}
\rowcolors{2}{lightgray}{white}
\begin{tabular}{lcccccccccc}
\toprule
\textbf{Method} &
Resolution & 
Dataset &
Train &
Abs Rel &
Sq Rel &
RMSE &
RMSE log &
$\delta < 1.25$ &
$\delta < 1.25^2$ &
$\delta < 1.25^3$\vspace{0.5mm}\\
\midrule
Garg et al.\cite{garg2016unsupervised} cap 50m & 620 x 188 & K & M & 0.169 & 1.080 &
5.104 & 0.273 & 0.740 & 0.904 & 0.962\\

Godard et al.~\cite{godard2018digging} & 640 x 192 & K & M & 0.129 & 1.112 & 5.180 & 0.205 & 0.851 & 0.952 & 0.978\\

SfMLearner~\cite{zhou2017unsupervised} (w/o explainability) & 416 x 128 & K & M
& 
0.221 & 2.226 & 7.527 & 0.294 & 0.676 & 0.885 & 0.954\\
SfMLearner~\cite{zhou2017unsupervised} & 416 x 128 & K & M & 0.208 & 1.768 & 6.856 & 0.283 & 0.678 & 0.885 & 0.957\\
SfMLearner~\cite{zhou2017unsupervised} & 416 x 128 & CS+K & M & 0.198 & 1.836 & 6.565 & 0.275 & 0.718 & 0.901 & 0.960\\
GeoNet~\cite{yin2018geonet} & 416 x 128 & K & M & 0.155 & 1.296 & 5.857 & 0.233 & 0.793 & 0.931 & 0.973 \\
GeoNet~\cite{yin2018geonet} & 416 x 128 & CS+K & M & 0.153 & 1.328 & 5.737 & 0.232 & 0.802 & 0.934 & 0.972 \\
Vid2Depth~\cite{mahjourian2018unsupervised} & 416 x 128 & K & M & 0.163 & 1.240 & 6.220 & 0.250 & 0.762 & 0.916 & 0.968 \\
Vid2Depth~\cite{mahjourian2018unsupervised} & 416 x 128 & CS+K & M & 0.159 & 1.231 & 5.912 & 0.243 & 0.784 & 0.923 & 0.970 \\

\midrule
UnDeepVO~\cite{li2017undeepvo} & 416 x 128 & K & S & 0.183 & 1.73 & 6.57 & 0.268 & - & - & - \\
Godard et al.~\cite{godard2017unsupervised} & 640 x 192 & K & S & 0.148 & 1.344 & 5.927 & 0.247 & 0.803 & 0.922 & 0.964\\
Godard et al.~\cite{godard2017unsupervised} & 640 x 192 & CS+K & S & 0.124 & 1.076 & 5.311 & 0.219 & 0.847 & 0.942 & 0.973\\
Godard et al.~\cite{godard2018digging} & 640 x 192 & K & S & 0.115 & 1.010 & 5.164 & 0.212 & \textbf{0.858} & 0.946 & 0.974\\
\textbf{Ours} & 1024 x 384 & K & S & 0.116 & 0.935 &  5.158 & 0.210 &  0.842 &  0.945 &  0.977\\
\textbf{Ours-SP} & 1024 x 384 & K & S & \textbf{0.112} & 0.880 & 4.959 & \textbf{0.207} & 0.850 & 0.947 & 0.977\\
\textbf{Ours-FA} & 1024 x 384 & K & S & 0.115 & 0.922 & 5.031 & 0.206 & 0.850 & \textbf{0.948} & \textbf{0.978}\\
\textbf{Ours-SP+FA} & 1024 x 384 & K & S & \textbf{0.112} & \textbf{0.875} & \textbf{4.958} & \textbf{0.207} & 0.852 & 0.947 & 0.977\\
\bottomrule
\end{tabular}\\
}
\caption{Single-view depth estimation results on the KITTI dataset~\cite{geiger2013vision} using the Eigen Split~\cite{eigen2014depth} for depths reported less than 80m, as indicated in~\cite{eigen2014depth}. The mode of self-supervision employed during training is reported under the \textbf{Train} column - Stereo (S), Mono (M). Above, we compare our baseline approach \textbf{(Ours)} along with the proposed sub-pixel convolutions variant \textbf{(Ours-SP)}. Training datasets used by previous methods are listed as either CS=Cityscapes~\cite{cordts2016cityscapes}, K=KITTI\cite{geiger2013vision}. For Abs Rel, Sq Rel, RMSE, and RMSE log, lower is better. For $\delta < 1.25$, $\delta < 1.25^2$ and $\delta < 1.25^3$, higher is better.}
\label{table:accuracy}
\vspace{-6mm}
\end{table*}

\subsubsection{Improving Disparity Estimation with Differentiable Flip-Augmentation Fine-Tuning}
In their previous works Godard et. al~\cite{godard2017unsupervised,godard2018digging} use a hand-engineered post-processing step to fuse the disparity estimates of the left image and the horizontally flipped image. While this reduces the artifacts at the borders of the image, we show that this technique can be used in a differentiable manner to allow further fine-tuning of the disparity network in an end-to-end manner. With the differentiable flip-augmentation training, we improve the baseline (Ours) and the sub-pixel variant (Ours-SP) on all metrics except the Abs. Rel which remains unchanged. Finally, by training with the subpixel-variant (Ours-SP) and fine-tuning with the flip-augmentation (Ours-FA) we are able to achieve state-of-the-art performance on the KITTI Eigen split benchmark as listed in Table~\ref{table:accuracy}.

\textbf{Effects of fine-tuning and pre-training}~Many recent state-of-the-art results~\cite{godard2018digging,guo2018learning} provide strong performance by either using pre-trained ImageNet weights~\cite{he2016deep} and fine-tuning or adapting the task domain from a model trained on an alternate dataset training. While we realize the implications of transferring well-conditioned model weights for warming up training, in this work we only consider the case of \textit{self-supervised training from scratch}. Despite training \textit{from scratch}, we show in Table~\ref{table:accuracy} that the performance of our models (Ours, Ours-SP, Ours-SP+FA) are competitive with those of recent state-of-the-art self-supervised disparity estimation methods~\cite{godard2017unsupervised,godard2018digging,guo2018learning} that utilize ImageNet pre-trained weights. 

\begin{figure}[!h]
  \centering
  {\renewcommand{\arraystretch}{0.4} 
   {\setlength{\tabcolsep}{0.2mm}
    \begin{tabular}{cc}
    \includegraphics[width=0.49\columnwidth]{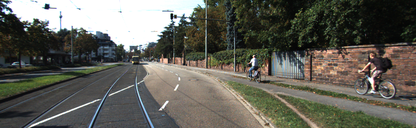}&
    \includegraphics[width=0.49\columnwidth]{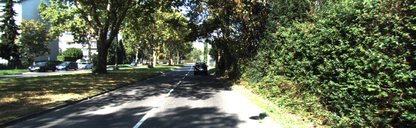}\\
    \includegraphics[width=0.49\columnwidth]{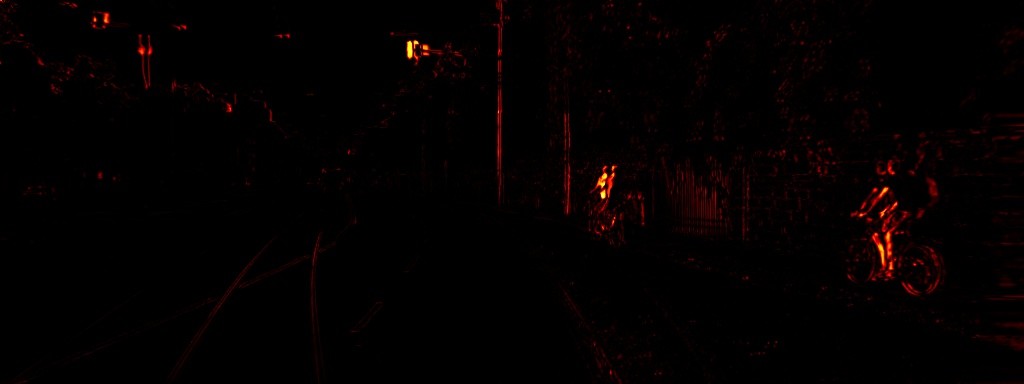}&
    \includegraphics[width=0.49\columnwidth]{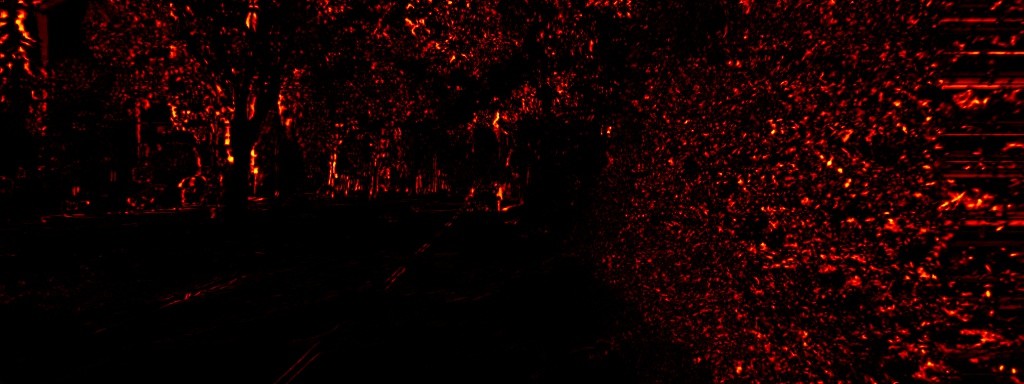} \\    
    \includegraphics[width=0.49\columnwidth]{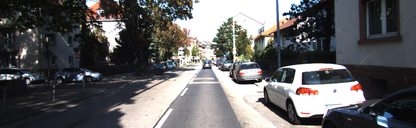}&
    \includegraphics[width=0.49\columnwidth]{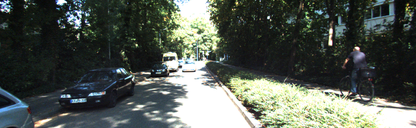} \\
    \includegraphics[width=0.49\columnwidth]{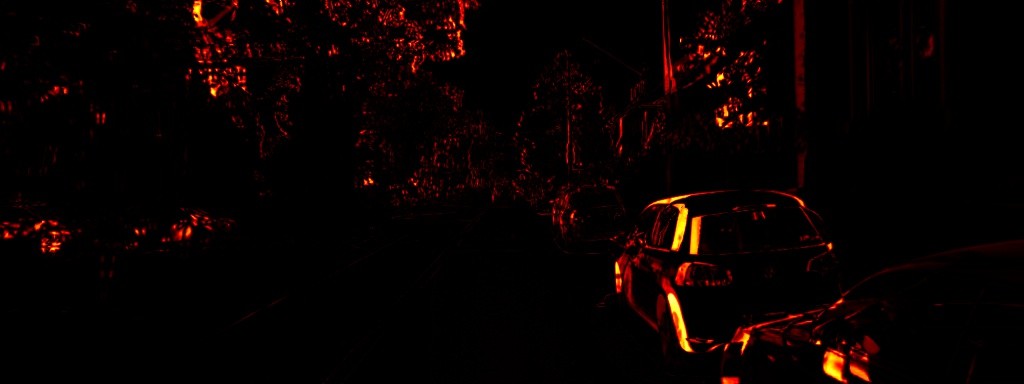}&
    \includegraphics[width=0.49\columnwidth]{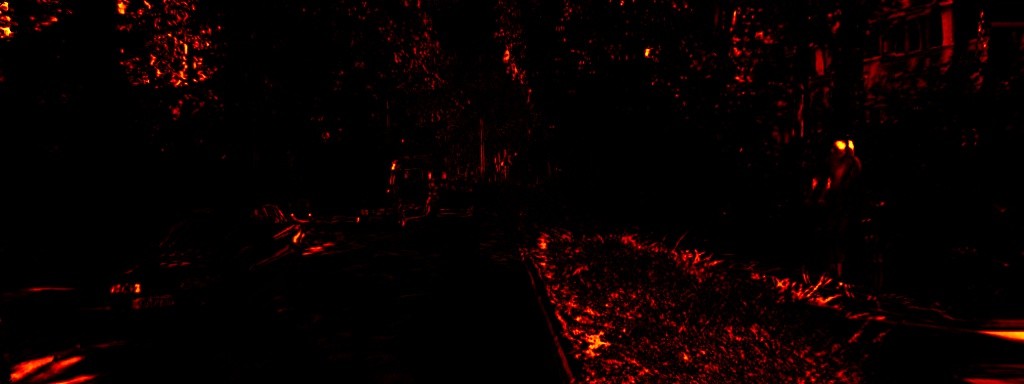}\\
  \end{tabular}}}
  \caption{Examples of pixel-wise photometric errors when reconstructing the \textit{right} image from the \textit{left} input image.}
  \label{fig:photometric}
\vspace{-3mm}
\end{figure}

\textbf{Qualitative results}~We contrast the results of our method alongside related methods in Figure~\ref{fig:disparity-illustration2}. We note that our method is able to capture with higher fidelity the sharpness of objects as compared to the state-of-the-art. The effect of our \textit{sub-pixel convolutions} is particularly noticeable around smaller objects (e.g. poles, traffic signs), where the super-resolved depths successfully recover the underlying geometry. Fig.~\ref{fig:photometric} shows examples of pixel-wise photometric errors induced when reconstructing the right image from the input left image.

\subsection{Pose Estimation}
To further validate our contributions, we perform a second set of experiments where we use our disparity network trained on stereo data to train a network which estimates the 6 DoF pose between subsequent monocular frames. Specifically, we are interested in recovering long-term trajectories that are metrically accurate and free of drift. Additionally, we bootstrap the training process with our disparity network, thus ensuring the trajectories are estimated with the correct scale factor.  

To estimate pose, we follow the architecture of~\cite{zhou2017unsupervised} \textit{without} the explainability mask layer. The network is fed the concatenation of a target image $I_{t}$ and a set of context images $I_{S}$, which are temporally adjacent to the target image. The network outputs the 6 DoF transformations between $I_t$ and the images in $I_{S}$ via the final $1\times 1$ convolutional layer. Following~\cite{brahmbhatt2018geometry,clark2017vinet,grassia1998practical}, we use the logarithm of a unit quaternion to parameterize the rotation in $\mathbb{R}^3$ and do not require an added normalization constraint unlike previous works~\cite{kendall2015posenet}. Finally, we use the logarithm and exponential maps to convert between a unit quaternion  and its log form~\cite{grassia1998practical}.

Formally, the network recovers a function $f_{\mathbf{x}}: (I_t,I_S) \to \mathbf{x}_{t \to s} = \begin{psmallmatrix}R & t\\ 0 & 1\end{psmallmatrix} \in SE\left(3\right)$, for all $s \in S$, where $\mathbf{x}_{t \to s}$  is the 6 DoF transformation between image $I_t$ and $I_s$. We train the pose network through an additional photometric loss between the target image $I_t$ and image $\hat{I_t}$ inferred via the mapping $\mathbf{x}_{t\to s}$ from the context image $I_s$. 
\begin{align}
      \mathcal{L}_{pm}(I_t,\hat{I_t}) = \alpha_2~\frac{1 - \text{SSIM}(I_t,\hat{I_t})}{2} + (1-\alpha_2)~\| I_t - \hat{I_t} \|
    \label{eq:photometric_loss_pose}   
\end{align}
We note here that, although similar to the $\mathcal{L}_{p}$ loss defined in Eq.~\ref{eq:loss-photo}, the multi-view photometric loss, $\mathcal{L}_{pm}$,  uses a different weight, $\alpha_2$, to trade-off between the $L1$ and the $SSIM$ components. In all our experiments, $\alpha_2=0.05$, thus the optimization favors the $L1$ component of the loss while training the pose network. This is important, as the $SSIM$ loss is better suited for images that are fronto-parallel (e.g. stereo camera images), an assumption which is often invalidated in images which are acquired sequentially as the camera is undergoing ego-motion. Furthermore, we jointly optimize Eq.~\ref{eq:overall-loss-disp} and Eq.~\ref{eq:photometric_loss_pose}, thus ensuring that the network which estimates disparity, $f_d$, does not diverge during this optimization step; this is important for recovering trajectories that are metrically accurate.   




For long term trajectory estimation we report Average Translational ($t_{rel}$) and Rotational ($r_{rel}$) RMS drift over trajectories of 100-800 meters. We use the KITTI odometry benchmark for evaluation, and specifically sequences 00 - 10, for which ground truth is available. We note that in this case we still train our disparity and pose networks on the KITTI \textit{Eigen} train split, with the mention that this data split includes all the images from sequences 01, 02, 06, 08, 09 and 10. We report our results on all sequences 00 - 10 in~\ref{table:tajectory-accuracy}, where we clearly mark the sequences that are used for training and testing, both for our method and the related work. We leave out model based methods (e.g.~\cite{mur2017orb,yang2018deep}) and limit our quantitative comparison to self-supervised learning based methods which are similar in nature to our approach. In all our experiments we use a context of size 3 (i.e. target frame plus 2 additional frames).

\begin{table}[!t]
\centering
{
\footnotesize
\setlength{\tabcolsep}{1.0em}
\begin{tabular}{l@{\hskip 1.5em}cc@{\hskip 1.5em}cc@{\hskip 1.5em}cc@{\hskip 1.5em}cc}


 & \multicolumn{2}{c}{SfMLearner~\cite{zhou2017unsupervised}$\ddagger$} & \multicolumn{2}{c}{UnDeepVO~\cite{li2017undeepvo}}  & \multicolumn{2}{c}{Ours} \\
\toprule
\textbf{Seq} & t$_{rel}$ & r$_{rel}$ & t$_{rel}$ & r$_{rel}$ & t$_{rel}$ & r$_{rel}$ \\
\midrule
00$^\dagger$  & 66.35 &  6.13 & 4.41 &  1.92 &  6.12 &  2.72  \\
03$^\dagger$  & 10.78 &  3.92 & 5.00 &  6.17 &  7.90 &  4.30  \\
04$^\dagger$  &  4.49 &  5.24 & 4.49 &  2.13 & 11.80 &  1.90  \\
05$^\dagger$  & 18.67 &  4.10 & 3.40 &  1.50 &  4.58 &  1.67  \\
07$^\dagger$  & 21.33 &  6.65 & 3.15 &  2.48 &  7.60 &  5.17  \\
\midrule
01$^*$        & 35.17 &  2.74 &69.07 &  1.60 & 13.48 &1.97  \\
02$^*$        & 58.75 &  3.58 & 5.58 &  2.44 & 3.48 & 1.10  \\
06$^*$        & 25.88 &  4.80 & 6.20 &  1.98 & 1.81 & 0.78  \\
08$^*$        & 21.90 &  2.91 & 4.08 &  1.79 & 2.25 & 0.84  \\
09$^*$        & 18.77 &  3.21 & 7.01 &  3.61 & 3.74 &  1.19  \\
10$^*$        & 14.33 &  3.30 &10.63 &  4.65 & 2.26 &  1.03\\
\midrule
\textbf{Train}   & 29.26 &  4.45 & 11.70 & 2.75 & \textbf{4.50} & \textbf{1.15} \\
\textbf{Test}   & 16.56 &  3.26 & 8.82 & 4.13 & \textbf{7.60} & \textbf{3.15}  \\
\textbf{Avg}   & 29.95 &  4.23 & 11.18 & 2.55 & \textbf{5.91} & \textbf{2.06}  \\
\bottomrule\end{tabular}\\\vspace{1mm}}
\caption{Long term trajectory results on the KITTI odometry benchmark. We report the following metrics: $t_{rel}$ - average translational RMSE drift (\%) on trajectories of length 100-800m, and $r_{rel}$ - average rotational RMSE drift ($^{{\circ}}/100m$) on trajectories of length 100-800m. $*$ and $\dagger$ represent train and respectively test seq. for our method. The methods of~\cite{zhou2017unsupervised} and~\cite{li2017undeepvo} are trained on seq. 00-08. $\ddagger$ The results of~\cite{zhou2017unsupervised} were scaled using the scale from the ground truth depth.}
\label{table:tajectory-accuracy}
\vspace{0mm}
\end{table}

\begin{figure}[t]
  \centering
  {\renewcommand{\arraystretch}{0.4} 
   {\setlength{\tabcolsep}{0.2mm}
    \begin{tabular}{ccc}
    \includegraphics[width=0.45\columnwidth]{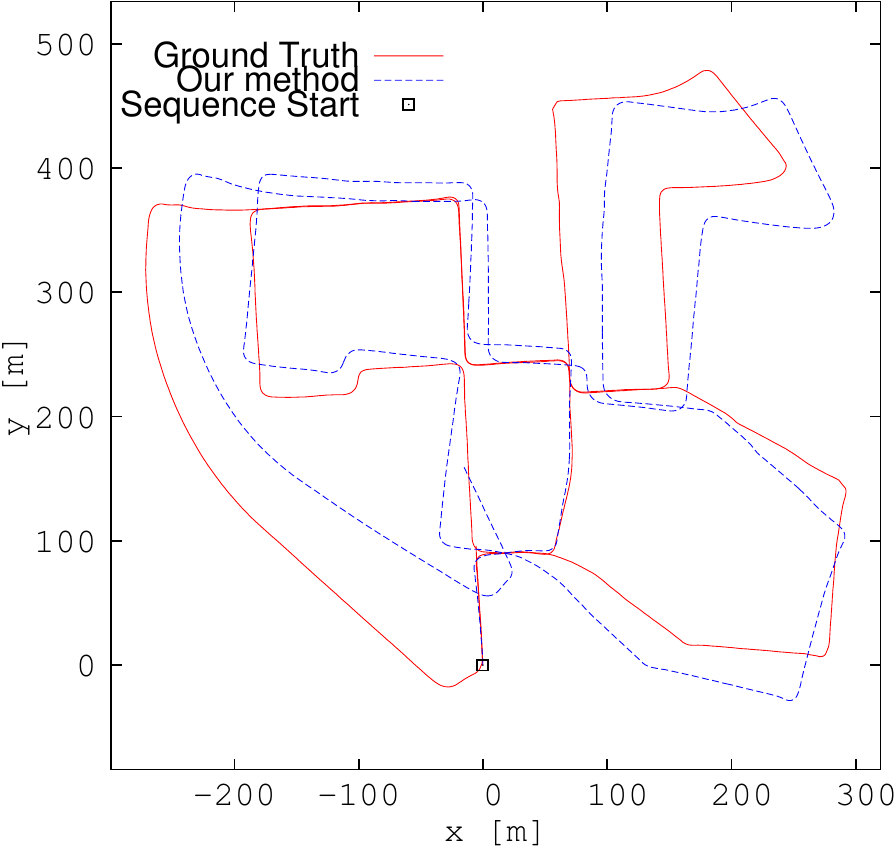}&
    \includegraphics[width=0.44\columnwidth]{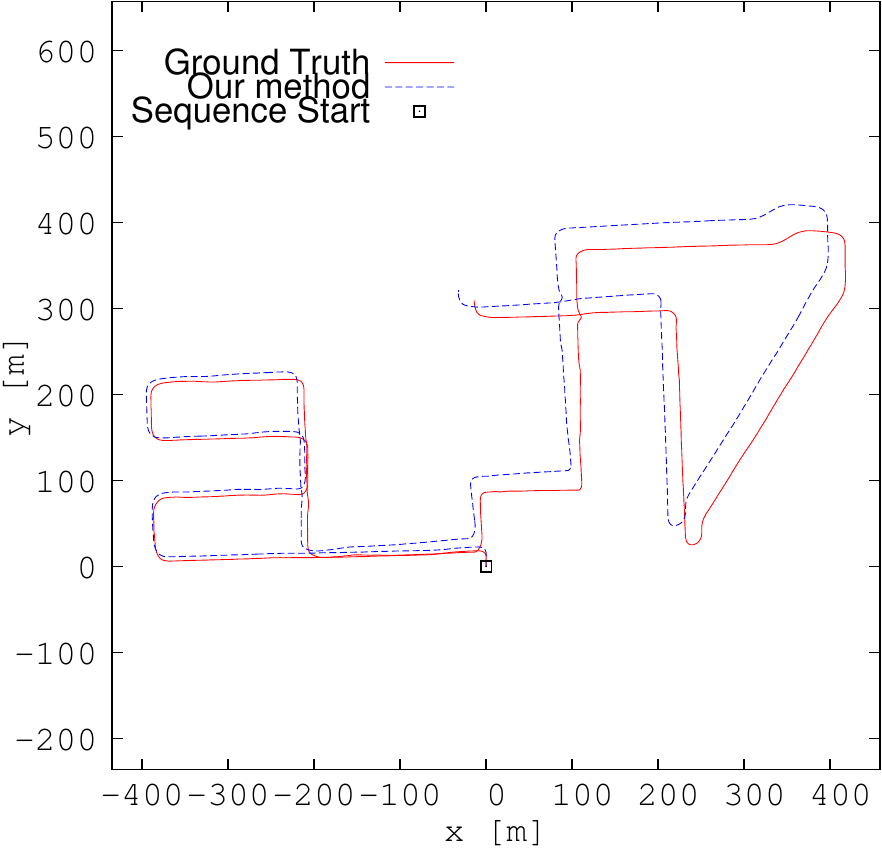}&
  \end{tabular}}}
  \caption{Illustrations of our pose estimation results on the KITTI odometry benchmark, sequences 00 (left) and 08 (right). Our results are rendered in blue while the ground truth is rendered in red.}
  \label{fig:trajectory-illustration}
\end{figure}

We compare against: (a) SfMLearner~\cite{zhou2017unsupervised} which is trained using monocular video and thus we scale their depth predictions using the scale from the ground truth; and (b) UnDeepVO~\cite{li2017undeepvo} which, like us, is trained on a combination of monocular and stereo imagery and returns metrically accurate depths and trajectories. We note that our quantitative results are superior to those of~\cite{li2017undeepvo}, which we attribute to the fact that our pose network is bootstrapped with much more accurate depth estimates. We further note that through the proposed combination of monocular and stereo losses our approach is able to overcome the scale ambiguity and recover metrically accurate trajectories which exhibit little drift over extended periods of time (see Table.~\ref{table:tajectory-accuracy} and Fig.~\ref{fig:trajectory-illustration}).


\subsection{Implementation}
We follow the implementation of~\cite{godard2017unsupervised} closely, and implement our depth estimation network in PyTorch. The sub-pixel convolution and differentiable flip-augmentation take advantage of the native \texttt{PixelShuffle} and \texttt{index\_select} operations in PyTorch, with the model and losses parallelized across 8 Titan V100s during training. We train the disparity network for 200 epochs using the Adam optimizer~\cite{kingma2014adam}. The learning rate and batch size are estimated via hyper-parameter search. In most cases, we use a batch size of 4 or 8, with an initial learning rate of 5e-4. As training proceeds, the learning rate is decayed every 40 epochs by a factor of 2. We set the following parameter values for all training runs: $\lambda_1=0.1$, $\lambda_2=0.01$, $\alpha=0.85$. For fine-tuning with the differentiable flip-augmentation layer, we use a learning rate of 5e-5, batch size of 2, and only consider the first 2 pyramid scales for computing the loss as the lower-resolution pyramid scales tend to over-regularize the depth maps.

\section{Conclusion} 
In this work, we propose two key extensions to self-supervised monocular disparity estimation that enables state-of-the-art performance on the public KITTI disparity estimation benchmark. Inspired by the strong performance in monocular disparity estimation in high-resolution regimes, we incorporate the concept of sub-pixel convolutions within a disparity estimation network to enable super-resolved depths. The super-resolved depths operating at higher-resolutions tend to reduce ambiguities in the self-supervised photometric loss estimation (unlike their lower-resolution counterparts), thereby resulting in improved monodepth estimation. In addition to super-resolution, we introduce a differentiable flip-augmentation layer that further reduces artifacts and ambiguities while training the monodepth model. Through experiments, we show that both contributions provide significant performance gains to the proposed self-supervised technique, resulting in state-of-the-art performance in depth estimation on the public KITTI benchmark. As a consequence of improved disparity estimation, we study its relation to the strongly correlated problem of pose estimation and show strong quantitative and qualitative performance compared to previous self-supervised pose estimation methods. 

\section*{Acknowledgments}
We would like to thank John Leonard, Mike Garrison, and the whole TRI-ML team for their support. Special thanks to Vitor Guizilini for his~help.
\balance
\bibliographystyle{templates/icra/bst/IEEEtran}
\bibliography{tex/references}

\end{document}